\documentclass[letterpaper, 10 pt, conference]{ieeeconf}  
\usepackage{amsfonts}
\usepackage{graphicx}
\usepackage{epstopdf}
\usepackage{booktabs}
\usepackage[colorlinks=true, allcolors=black]{hyperref}
\usepackage{array, caption, threeparttable}
\usepackage[font=small,labelfont=rm,labelsep=none]{caption}
\usepackage{graphicx,times,psfig,amsmath} 
\usepackage{epsfig,amssymb,latexsym,psfrag,graphicx}
\usepackage{setspace, color}
\usepackage{verbatim}
\usepackage{subfigure}
\usepackage{cite}
\usepackage{mwe}
\usepackage{float}
\usepackage{caption}
\usepackage{algorithm}
\usepackage{algpseudocode}
\usepackage{balance}
\usepackage{tikz}
\newtheorem{theorem}{Theorem}

\newtheorem{lemma}{Lemma}
\newtheorem{proposition}{Proposition}

\newtheorem{assumption}{Assumption}
\newtheorem{definition}{Definition}
\usetikzlibrary{circuits.ee.IEC}
\usetikzlibrary{angles,quotes} 
\usetikzlibrary{arrows,arrows.meta}
\usetikzlibrary{shapes.geometric}
\usetikzlibrary{positioning,calc,patterns,decorations.pathmorphing,decorations.markings}
\usepackage{setspace}
\newcommand{\bea}{\begin{eqnarray}}  \newcommand{\eea}{\end{eqnarray}}
\newcommand{\be}{\begin{equation}}   \newcommand{\ee}{\end{equation}}
\newcommand{\ba}{\begin{array}}      \newcommand{\ea}{\end{array}}
            
\newcommand{\nnum}{\nonumber}

      \def\QED{~\rule[-1pt]{5pt}{5pt}\par\medskip}

\newcommand{\bd}{\begin{definition}} \newcommand{\ed}{\end{definition}}
\newcommand{\bt}{\begin{theorem}}    \newcommand{\et}{\end{theorem}}

\newcommand{\bl}{\begin{lemma}}      \newcommand{\el}{\end{lemma}}

\IEEEoverridecommandlockouts                              

\title{\LARGE \bf
Real-Time Adaptive Safety-Critical Control with Gaussian Processes in High-Order Uncertain Models
}

\author{Yu Zhang$^{1}$, Long Wen$^{1}$, Xiangtong Yao$^{1}$, Zhenshan Bing$^{1*}$, Linghuan Kong$^{2}$, \\Wei He$^{3}$, and Alois Knoll$^{1}$
\thanks{$^{1}$Y. Zhang, L. Wen, X. Yao, Z. Bing, and A. Knoll are with the Department of Informatics, Technical University of Munich, Germany.
Corresponding author: Zhenshan Bing ({\tt\small zhenshan.bing@tum.de}).}%
\thanks{$^{2}$L. Kong is with the Department of Electrical and Computer Engineering, Faculty of Science and Technology, University of Macau, Macau, China.
        }%
\thanks{$^{3}$W. He is with the School of Intelligence Science and Technology, University of Science and Technology Beijing, Beijing 100083, China, and also with the Key Laboratory of Intelligent Bionic Unmanned Systems, Ministry of Education, University of Science and Technology Beijing, Beijing 100083, China.
        }%
}

\begin{document}

\maketitle
\thispagestyle{empty}
\pagestyle{empty}

\begin{abstract}

This paper presents an adaptive online learning framework for systems with uncertain parameters to ensure safety-critical control in non-stationary environments. Our approach consists of two phases. The initial phase is centered on a novel sparse Gaussian process (GP) framework. We first integrate a forgetting factor to refine a variational sparse GP algorithm, thus enhancing its adaptability. Subsequently, the hyperparameters of the Gaussian model are trained with a specially compound kernel, and the Gaussian model's online inferential capability and computational efficiency are strengthened by updating a solitary inducing point derived from newly samples, in conjunction with the learned hyperparameters. In the second phase, we propose a safety filter based on high order control barrier functions (HOCBFs), synergized with the previously trained learning model. By leveraging the compound kernel from the first phase, we effectively address the inherent limitations of GPs in handling high-dimensional problems for real-time applications.
The derived controller ensures a rigorous lower bound on the probability of satisfying the safety specification. Finally, the efficacy of our proposed algorithm is demonstrated through real-time obstacle avoidance experiments executed using both simulation platform and a real-world 7-DOF robot.

\end{abstract}

\section{INTRODUCTION}

Recently, the increasing demand for safety-certified controllers in autonomous systems has been driven by the emergence of control barrier functions (CBFs). CBFs offer a framework for ensuring state forward-invariance, requiring just a single-step horizon for assurance \cite{8796030}. This reduces computational effort, making it suitable for real-time applications, such as walking robot \cite{nguyen2015safety}, autonomous driving \cite{9029446}, robotic manipulator\cite{10132404}, drone control \cite{8460471}, and multi-agent system \cite{9112342}. Note that CBFs are often integrated into a quadratic programming (QP) problem as constraints, acting as safety filters to ensure system safety. This leads to a strong dependence on precise models, which are difficult to obtain in real-world applications \cite{10341769}. Therefore, ensuring system safety in non-stationary environments with imprecise system models has become a recent research focus area.
\begin{figure}[htbp]
    \centering
    \includegraphics[width=3.43in,height=2.28in]{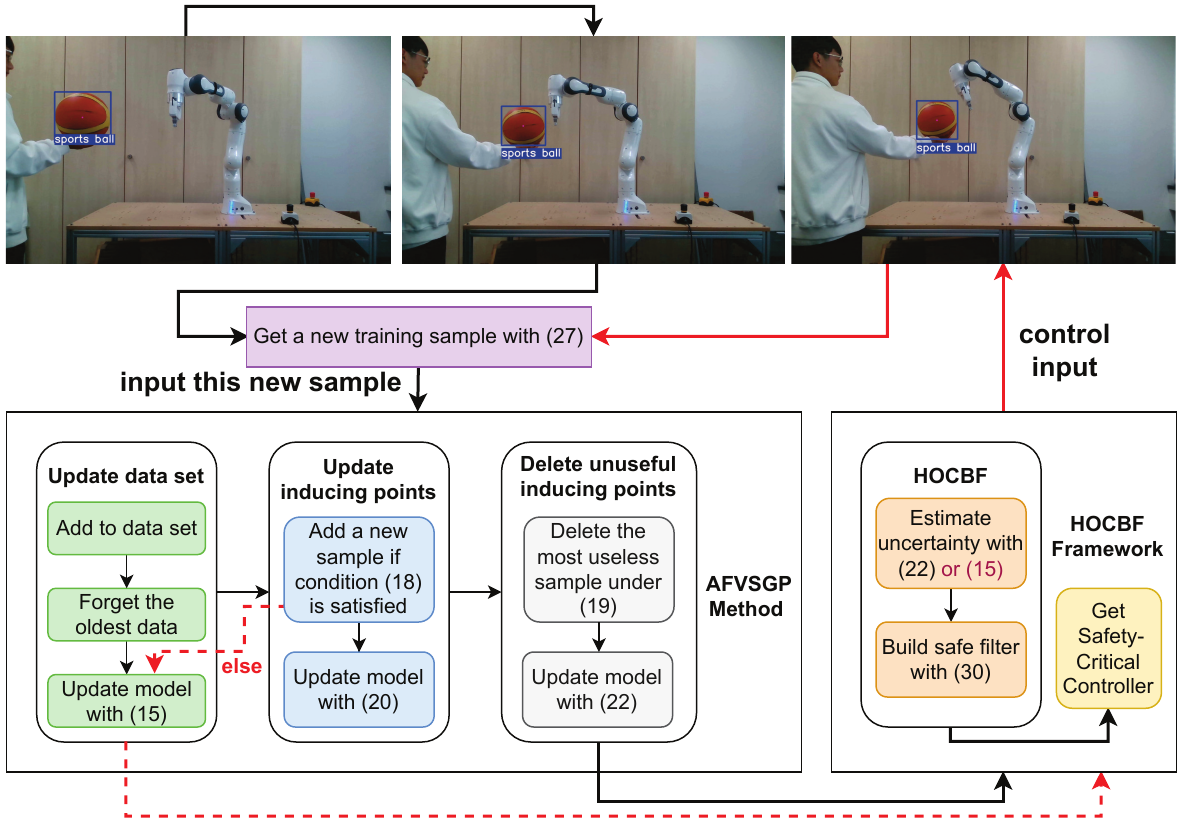}
    \caption{\quad An overview of the proposed AFVSGP-HOCBF framework. AFVSGP estimates the model-induced uncertainty affecting HOCBF, while HOCBF is applied to guarantee system safety. }
    \label{fig1}
    \vspace{-0.8em}
\end{figure}

System dynamics are typically viewed as a combination of the nominal and an uncertain component. This uncertainty arises from inaccuracies in the model parameters, and models with this kind of uncertainty are referred to as uncertain models. In scenarios where the nominal part closely approximates the real model, robust CBFs are suitable to be employed to compensate for the discrepancies \cite{9147463,9353988,9129764,dong2023novel,9772990,beckers2017stable,10160626,9466373}. However, when the disparity between the two becomes overly large, this method tends to be excessively conservative. For example, the robotic manipulator starts avoiding obstacles even when they are far away, significantly reducing its operational workspace for its inherent tasks. To address the issue of excessive conservatism, the focus has increasingly shifted towards learning-based methods. Among these, Gaussian processes (GPs) stand out because they offer a Bayesian, data-driven approach that requires minimal prior knowledge for accurately modeling complex functions of arbitrary complexity \cite{kocijan2016modelling}. In \cite{9303847}, CBFs were firstly combined with GPs to devise controllers ensuring safety for affine systems with unknown drift term, and \cite{9196709} further made a comparison between GPs and dropout neural networks \cite{gal2016dropout}, as well as ALPaCA \cite{harrison2020meta}, highlighting GPs' superiority.

The early implementations of GP-CBFs for ensuring system safety exhibited several limitations. Firstly, offline learning strategies in GPs require extensive data collection to formulate a comprehensive global model, thereby reducing its adaptability in dynamic changes of the environment. Secondly, the inherent complexity of high-dimensional nonlinear models poses challenges for GPs, as a GP can only estimate a scalar function. An excessive number of GPs substantially diminishes the likelihood of maintaining system security. Lastly, introducing uncertainties in the input gain matrix of system dynamics violate the convexity in the QP problem, thereby increasing its solution time. Existing methods mostly address these drawbacks of GP-CBFs approaches through two primary strategies. The first strategy employs sparse online GP (SOGP) \cite{bui2017streaming}, to decrease the time complexity from $\mathcal{O}\left(N^3\right)$ to $\mathcal{O}\left(N M^2\right)$, where $N$ is the number of data and $M$ represents the number of most representative points in the data, i.e., $M \ll N$. The second strategy addresses the challenge of inferring control inputs by altering the kernel functions within the GPs. In \cite{9658123}, a novel matric variate GP was introduced to simultaneously estimate the drift term and input gain of affine systems. The accompanying time complexity is $\mathcal{O}\left(N^3\right)$ with learning hyperparameters and $\mathcal{O}\left(NM^2\right)$ without learning hyperparameters. In \cite{9683743}, a GPs framework based on a compound kernel function was applied  to ensure the GP-CBF-QP problem's convexity, providing both necessary and sufficient conditions for the feasible solution, with a time complexity of $\mathcal{O}\left(N^3\right)$.

However, to enable real-time safety in non-stationary environments using GP-CBFs control algorithms, three key challenges must still be addressed. Firstly, the algorithm should accommodate input gain uncertainties and be applicable to high-dimensional nonlinear models. Secondly, the adaptive selection of training data and inducing points must be considered to reflect the evolving nature of the environment. Lastly, optimizing time complexity in GPs is critical, especially for matrix inversion tasks during online model updates and inference. In this paper, we propose a novel adaptive online learning framework and the main contributions are outlined below. 1)  A specific kernel function is introduced in the variational sparse GP (VSGP), enhancing online inference in high-dimensional nonlinear models with uncertain input gains, while maintaining VSGP's inherent benefit of reducing overfitting. 2) This is the first effort to incorporate novel forgetting mechanisms and rules in the VSGP-CBFs methodology. The goal is to make the model more adaptive to environmental changes by selecting the most informationally-rich training and inducing points. 3) The online update of the model's posterior distribution reduces the time complexity to $\mathcal{O}\left(M^3\right)$, as opposed to $\mathcal{O}\left(NM^2\right)$ in the existing methods. 4) This work explores high order CBFs instead of conventional CBFs, providing a more comprehensive theoretical framework. The efficacy and real-time performance of the proposed algorithm were substantiated through dynamic obstacle avoidance tasks, executed both in the simulation platform and on a real-world 7-DOF robotic manipulator.

\section{Background \& Preliminaries}
In this section, we provide a brief introduction to the mathematical foundations of high order CBFs (HOCBF) \cite{xu2015robustness} and VSGP \cite{titsias2009variational}, upon which our novel adaptive online learning framework builts.
\subsection{High Order Control Barrier Function}
The paper focuses on a class of control-affine systems, as described below.
\begin{equation} \label{my_equ1}
\dot{\boldsymbol{x}}=f(\boldsymbol{x})+g(\boldsymbol{x}) \boldsymbol{u}
\end{equation}
where $\boldsymbol{x} \in \mathcal{X} \subset \mathbb{R}^{n}$ is the state; $\boldsymbol{u} \in \mathcal{U} \subset \mathbb{R}^{m}$ is the control input; $\mathcal{U}$ is a compact set; $f(\boldsymbol{x})$  and $g(\boldsymbol{x})$ are locally Lipschitz continuous and represent drift term and input gain, respectively. The manipulator $\mathcal{A}$ is controlled by a Lipschitz continuous control law $\boldsymbol{u} = k(\boldsymbol{x})$.

we consider a closed convex set $\mathcal{F}$ and represent it as the super level set of a continuously differentiable function $h: \mathcal{X} \subset \mathbb{R}^{n} \subset \mathbb{R}$, i.e.,
\begin{equation} \label{my_equ3}
\begin{aligned}
\mathcal{F} & =\left\{x \in \mathcal{X} \subset \mathbb{R}^n: h(x) \geq 0\right\} \\
\partial \mathcal{F} & =\left\{x \in \mathcal{X} \subset \mathbb{R}^n: h(x)=0\right\} \\
\operatorname{Int}(\mathcal{F}) & =\left\{x \in \mathcal{X} \subset \mathbb{R}^n: h(x)>0\right\}
\end{aligned}
\end{equation}
where $\mathcal{F}$ is the nonempty safe set, i.e., $\operatorname{Int}(\mathcal{F}) \neq \emptyset$, and we assume that there is no isolated point in $\mathcal{F}$.

\begin{definition}
 $(\text{Invariant Set})$
Assuming that the initial conditions $\boldsymbol{x}_{0} \in \mathcal{F} \subset \mathcal{X}$ and subsequent states $\boldsymbol{x} \in \mathcal{F} \subset \mathcal{X}$ for time $t \in\left[t_0, t_{\max }\right)$ can be achieved by designing an appropriate input $\boldsymbol{u}(t)$, we can assert that the set $\mathcal{F}$ is a controlled invariant set for system \eqref{my_equ1}. Additionally, system \eqref{my_equ1} is forward complete when $t_{\max }=\infty$.
\end{definition}

The \textit{Definition 1} dictates that for any state \( \boldsymbol{x} \in \mathcal{F} \), it can be sustained within the set \( \mathcal{F} \) using permissible control inputs. Furthermore, for system \eqref{my_equ1}, its forward completeness for \( t_{\max} = \infty \) suggests that, under the prescribed control input, the system's evolution can span an indefinite time horizon while staying inside the set \( \mathcal{F} \).

\begin{definition}
  \text {(Relative degree)} The relative degree of a control barrier function $h(\cdot): \mathbb{R}^{n} \rightarrow \mathbb{R}$ with respect to the system \eqref{my_equ1} is defined as the minimum number of times the function needs to be differentiated along the dynamics of \eqref{my_equ1} until the control input $\boldsymbol{u}$ appears explicitly.
\end{definition}

\begin{definition}
  (HOCBF)  A function $h(x): \mathbb{R}^n \rightarrow \mathbb{R}$ is a HOCBF of relative degree $m$ for system if there exist $m-i$th order differentiable class $\mathcal{K}$ functions $\alpha_i$ and a class $\mathcal{K}$ function $a_m$, such that: $\sup_{u \in U} [ L_f^m b(x) + L_g L_f^{m-1} b(x) u + S(b(x)) + \alpha_m(\psi_{m-1}(x))] \geq 0$ for all $x \in C_1 \cap \cdots \cap C_m$, where $C_i=\left\{x \in \mathbb{R}^n: \psi_{i-1}(x) \geq 0\right\}$ for all $i \in\{1, \ldots, m\}$, $\psi_i(x)=\dot{\psi}_{i-1}(x)+\alpha_i\left(\psi_{i-1}(x)\right)$, and $\psi_0(x)=h(x)$. $S\left(b_i(x)\right)$ denotes the remaining Lie derivative along $f$ with degree less than or equal to $m-1$, that is, $S\left(b_i(x)\right)=\sum_{i=1}^{m-1} L_f^i\left(\alpha_{m-i} \circ \psi_{m-i-1}\right)(x)$.
\end{definition}

\begin{proposition}
  \text {(Sufficient condition for safety)} Consider the zero super-level set of a continuously differentiable control barrier function $h(\boldsymbol{x})$, denoted as $\mathcal{F}$. If the condition $x_{0} \in C_1 \cap \cdots \cap C_m$ holds, then the set $C_1 \cap \cdots \cap C_m$ is forward invariant, and the system can be guaranteed to operate safely via the implementation of a control policy, denoted as $\pi(\mathbf{x})$, that satisfies the constraint $\pi(\mathbf{x}) \in \{\mathbf{u} \in \mathcal{U} \mid L_f^m b(x) + L_g L_f^{m-1} b(x) u + S(b(x))+\alpha_m(\psi_{m-1}(x))\geq 0\}.$
\end{proposition}

To identify a safety controller for system \eqref{my_equ1}, the HOCBF is finally integrated with a QP problem as delineated below to finds the minimum perturbation on $u_{\rm nom}$.
\begin{equation} \label{my_equ04}
\begin{aligned}
u(x)= \underset{u \in \mathbb{R}^m}{\operatorname{argmin}}\; & \frac{1}{2}\|u-u_{\rm nom}\|^2 \qquad \qquad  \textrm{(HOCBF-QP)} \\
\text { s.t. } & L_f^m b(x) + L_g L_f^{m-1} b(x) u + S(b(x))\\
& \qquad \qquad \qquad +\alpha_m(\psi_{m-1}(x))\geq 0
\end{aligned}
\end{equation}
where the term $u_{\rm nom}$ represents a nominal feedback controller. Note that $u_{\rm nom}$ can be derived through various methodologies including model predictive control, control Lyapunov function, neural network-based strategies, proportional derivative control, or reinforcement learning.
\tikzstyle{arrow} = [thick,->,>=stealth]
\begin{figure*}[!t]
  \centering
    \begin{tikzpicture}
        \begin{scope}[xshift=3cm, yshift=0.4cm]
            \node[above right] (fig73) at (0,0){\includegraphics[width=0.3\linewidth,height=0.15\linewidth]{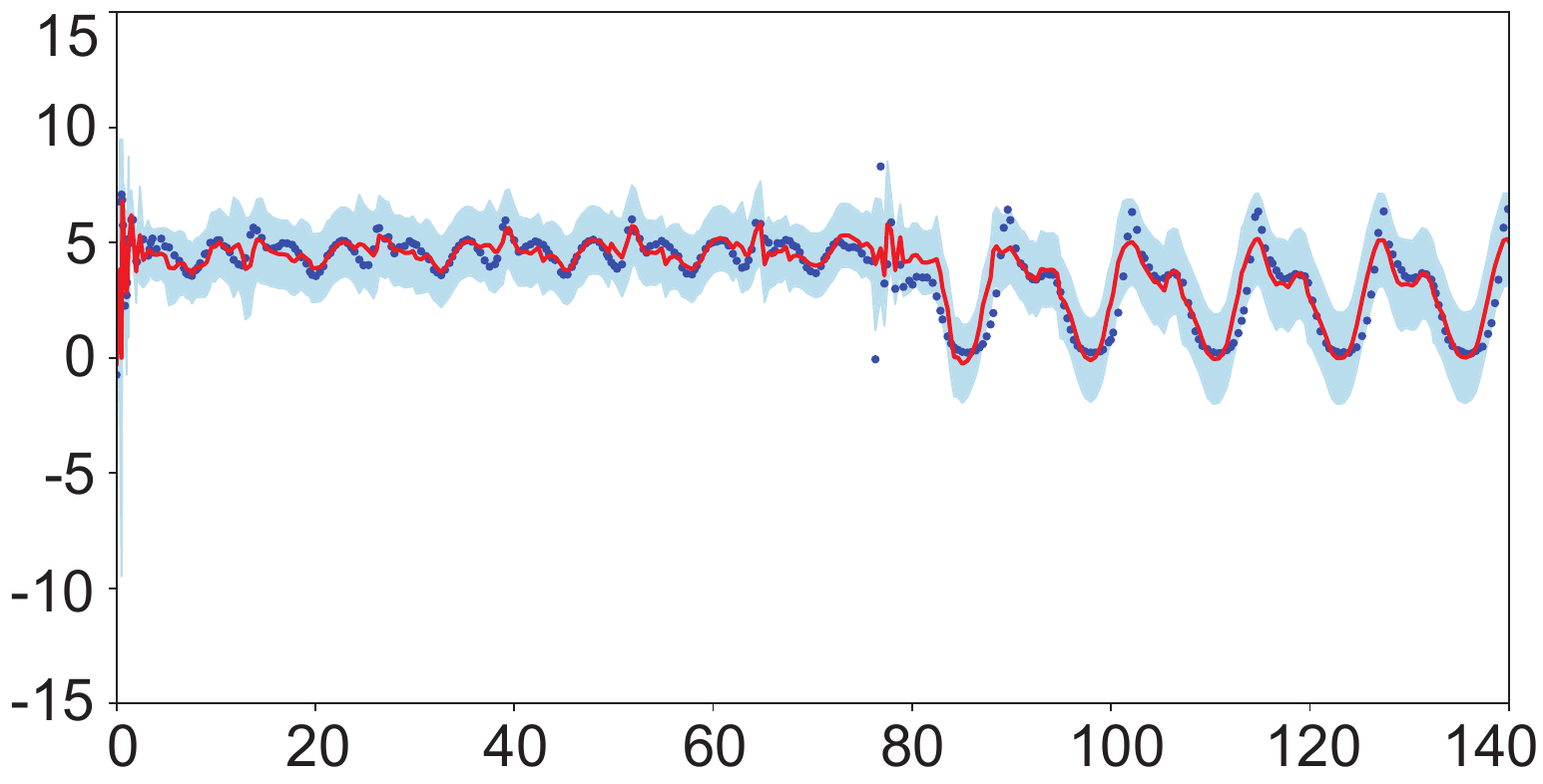}};
            \node at ($(fig73.south)+(0,-0.35)$) {\normalsize (c) static obstacle with SOGP};
            \node[rectangle, rounded corners=1pt,minimum width=2.1cm,minimum height=0.3cm,draw=gray,fill=white!10](box) at ($(fig73.north east)+(-1.35,-0.35)$) {};
            \node (label) at ($(box.west)+(0.9,0.03)$) {\tiny $\Delta B^{(m)}$};
            \node (label2) at ($(box.east)+(-0.2,0.03)$) {\tiny $m^*$};
            \draw[fill=blue] ($(label.west)+(-0.06,-0.05)$) circle (0.05);
            \draw[draw=red,line width=1pt] ($(label2.west)+(-0.25,-0.05)$) -- ($(label2.west)+(0.05,-0.05)$);
            \draw[draw=green!90!black,dashed,line width=1pt] ($(fig73.north)+(0.38,-0.2)$) -- ($(fig73.south)+(0.38,0.35)$);
            \node(text) at ($(fig73.south)+(-0.5,0.6)$) {\footnotesize switch motion};
            \draw[arrow,red] (text.north) -- ++(0.8,0.4);
            \node[rotate=90] at ($(fig73.west)+(0,0)$) {\scriptsize TUC$[m\slash s^2]$};
            \node at ($(fig73.south)+(0,-0.02)$) {\scriptsize time$[s]$};
        \end{scope}

        \begin{scope}[xshift=-3.1cm, yshift=0.4cm]
            \node[above right] (fig74) at (0,0){\includegraphics[width=0.308\linewidth,height=0.15\linewidth]{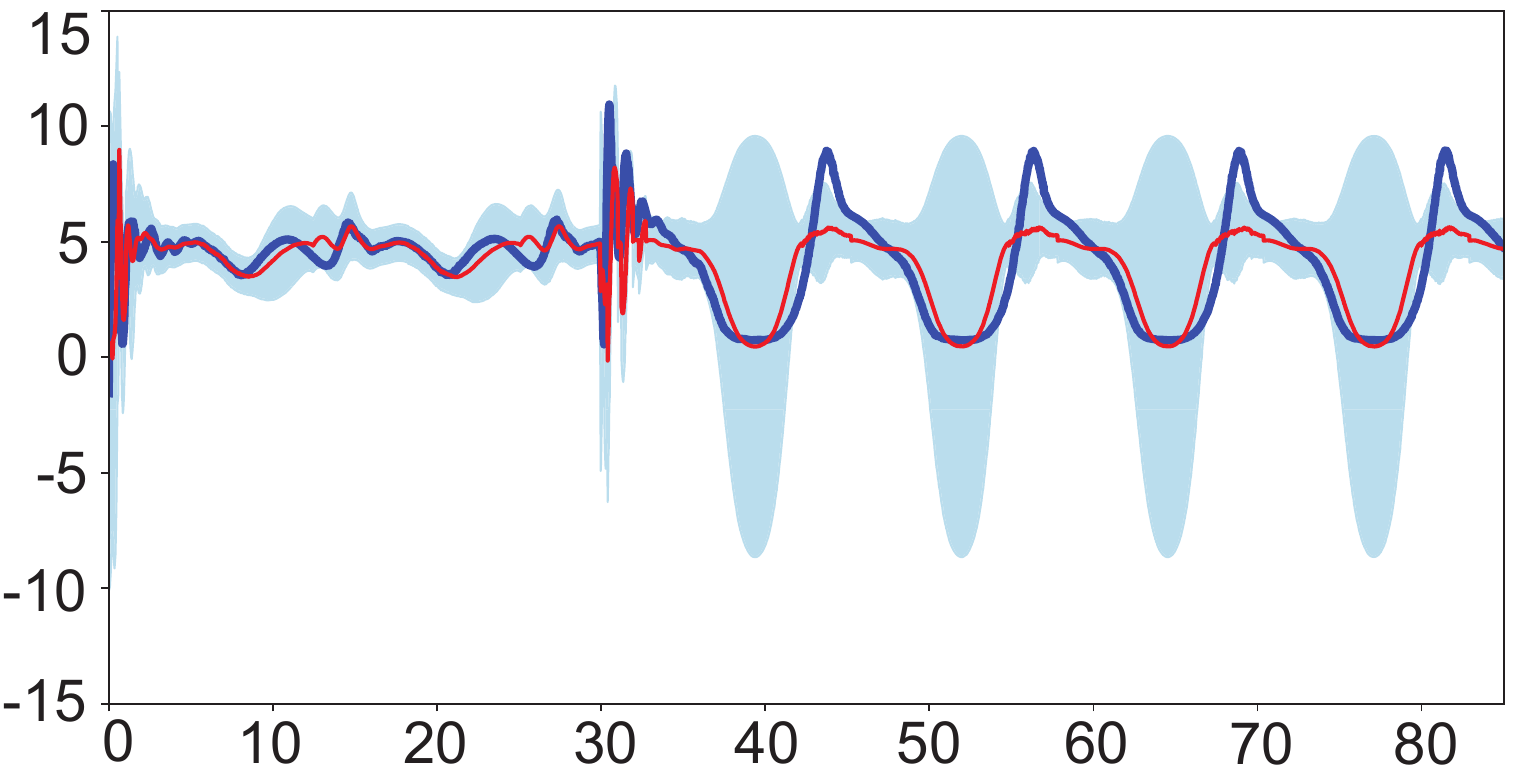}};
            \node at ($(fig74.south)+(0,-0.35)$) {\normalsize (b) static obstacle with VSGP};
            \node[rectangle, rounded corners=1pt,minimum width=2.1cm,minimum height=0.3cm,draw=gray,fill=white!10](box) at ($(fig74.north east)+(-1.25,-0.4)$) {};
            \node (label) at ($(box.west)+(0.9,0.03)$) {\tiny $\Delta B^{(m)}$};
            \node (label2) at ($(box.east)+(-0.2,0.03)$) {\tiny $m^*$};
            \draw[draw=blue,line width=1pt] ($(label.west)+(-0.25,-0.05)$) -- ($(label.west)+(0.05,-0.05)$) node (legend){};
            \draw[draw=red,line width=1pt] ($(label2.west)+(-0.25,-0.05)$) -- ($(label2.west)+(0.05,-0.05)$);
            \draw[draw=green!90!black,dashed,line width=1pt] ($(fig74.north)+(-0.52,-0.2)$) -- ($(fig74.south)+(-0.52,0.35)$);
            \node(text) at ($(fig74.south)+(-1.43,0.6)$) {\footnotesize switch motion};
            \draw[arrow,red] (text.north) -- ++(0.8,0.4);
            \node[rotate=90] at ($(fig74.west)+(0.1,0)$) {\scriptsize TUC$[m\slash s^2]$};
            \node at ($(fig74.south)+(0,-0.02)$) {\scriptsize time$[s]$};
        \end{scope}

        \begin{scope}[xshift=-9cm, yshift=0.4cm]
            \node[above right] (fig75) at (0,0){\includegraphics[width=0.3\linewidth,height=0.147\linewidth]{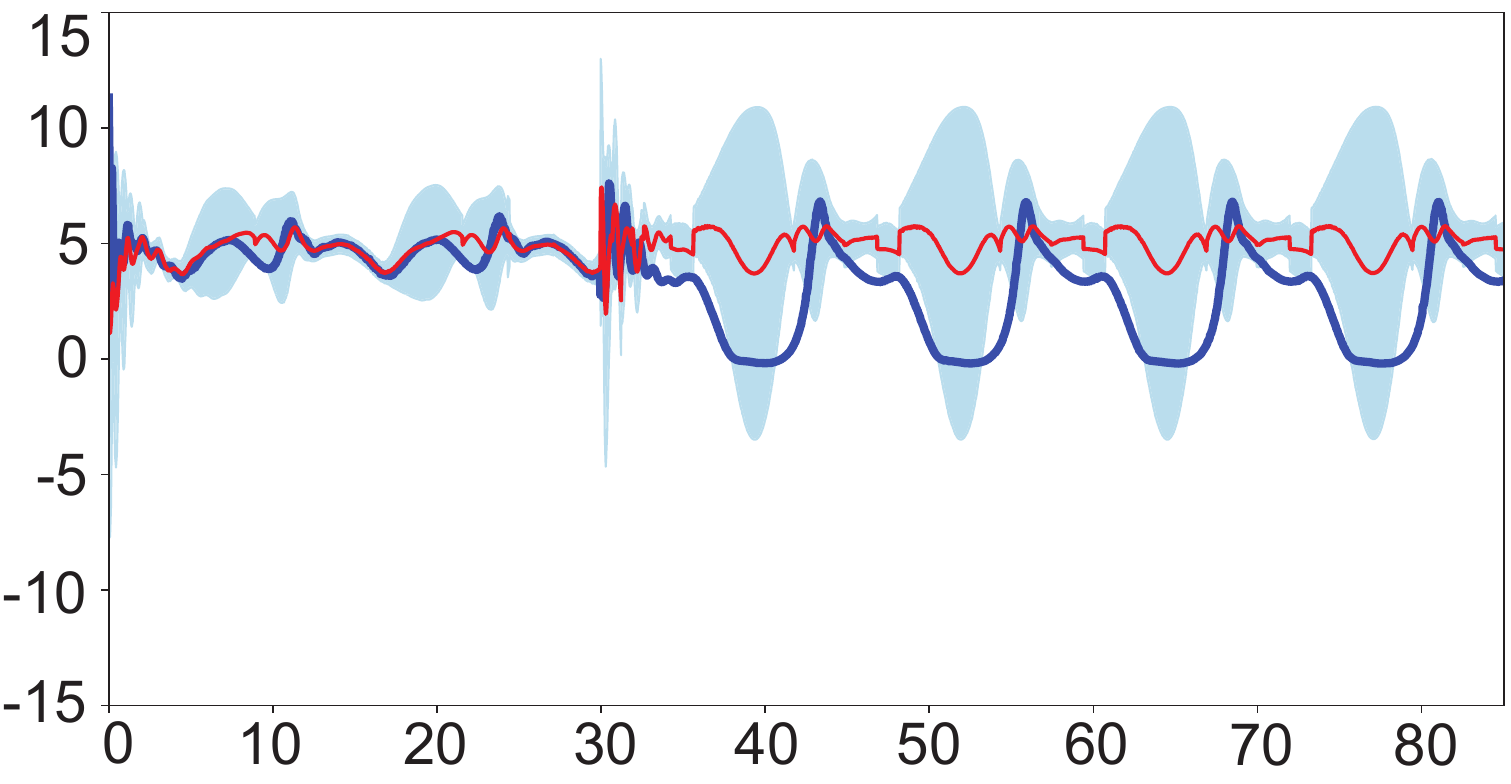}};
            \node at ($(fig75.south)+(0,-0.35)$) {\normalsize (a) static obstacle with GP};
            \node[rectangle, rounded corners=1pt,minimum width=2.1cm,minimum height=0.3cm,draw=gray,fill=white!10](box) at ($(fig75.north east)+(-1.25,-0.33)$) {};
            \node (label) at ($(box.west)+(0.9,0.03)$) {\tiny $\Delta B^{(m)}$};
            \node (label2) at ($(box.east)+(-0.2,0.03)$) {\tiny $m^*$};
            \draw[draw=blue,line width=1pt] ($(label.west)+(-0.25,-0.05)$) -- ($(label.west)+(0.05,-0.05)$) node (legend){};
            \draw[draw=red,line width=1pt] ($(label2.west)+(-0.25,-0.05)$) -- ($(label2.west)+(0.05,-0.05)$);
            \draw[draw=green!90!black,dashed,line width=1pt] ($(fig75.north)+(-0.52,-0.2)$) -- ($(fig75.south)+(-0.52,0.35)$);
            \node(text) at ($(fig75.south)+(-1.43,0.6)$) {\footnotesize switch motion};
            \draw[arrow,red] (text.north) -- ++(0.8,0.4);
            \node[rotate=90] at ($(fig75.west)+(0,0)$) {\scriptsize TUC$[m\slash s^2]$};
            \node at ($(fig75.south)+(0,-0.02)$) {\scriptsize time$[s]$};
        \end{scope}

        \begin{scope}[xshift=-9cm, yshift=-3cm]
            \node[above right] (fig76) at (0,0){\includegraphics[width=0.3\linewidth,height=0.15\linewidth]{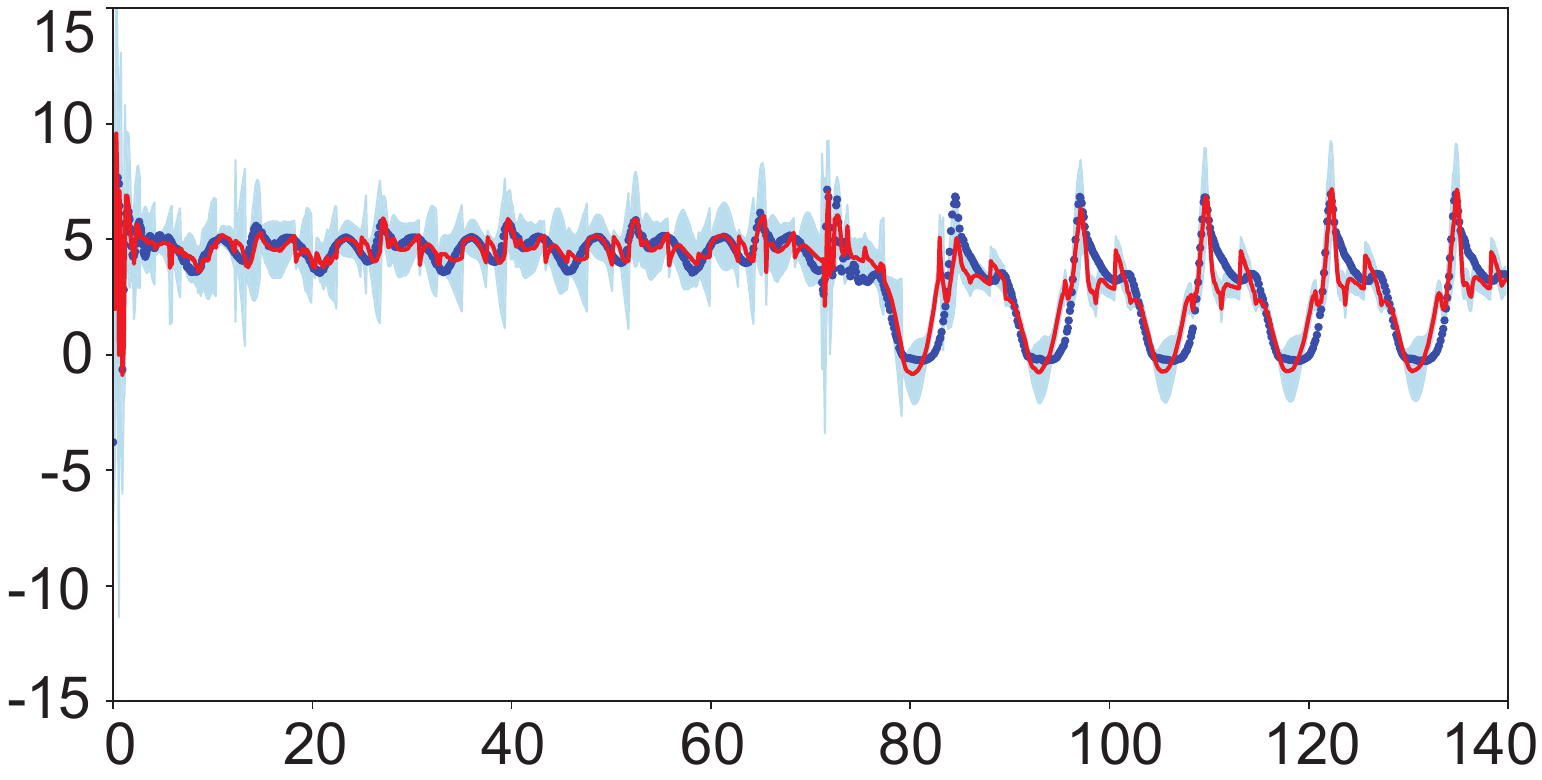}};
            \node at ($(fig76.south)+(0,-0.35)$) {\normalsize (d) static obstacle with AFVSGP (ours)};
            \node[rectangle, rounded corners=1pt,minimum width=2.1cm,minimum height=0.3cm,draw=gray,fill=white!10](box) at ($(fig76.north east)+(-1.35,-0.35)$) {};
            \node (label) at ($(box.west)+(0.9,0.03)$) {\tiny $\Delta B^{(m)}$};
            \node (label2) at ($(box.east)+(-0.2,0.03)$) {\tiny $m^*$};
            \draw[fill=blue] ($(label.west)+(-0.06,-0.05)$) circle (0.05);
            \draw[draw=green!90!black,dashed,line width=1pt] ($(fig76.north)+(0.22,-0.2)$) -- ($(fig76.south)+(0.22,0.35)$);
            \draw[draw=red,line width=1pt] ($(label2.west)+(-0.25,-0.05)$) -- ($(label2.west)+(0.05,-0.05)$);
            \node(text) at ($(fig76.south)+(-0.7,0.6)$) {\footnotesize switch motion};
            \draw[arrow,red] (text.north) -- ++(0.8,0.4);
            \node[rotate=90] at ($(fig76.west)+(0.,0)$) {\scriptsize TUC$[m\slash s^2]$};
            \node at ($(fig76.south)+(0,-0.02)$) {\scriptsize time$[s]$};
        \end{scope}

        \begin{scope}[xshift=-3cm, yshift=-3cm]
            \node[above right] (fig68) at (0,0){\includegraphics[width=0.3\linewidth,height=0.15\linewidth]{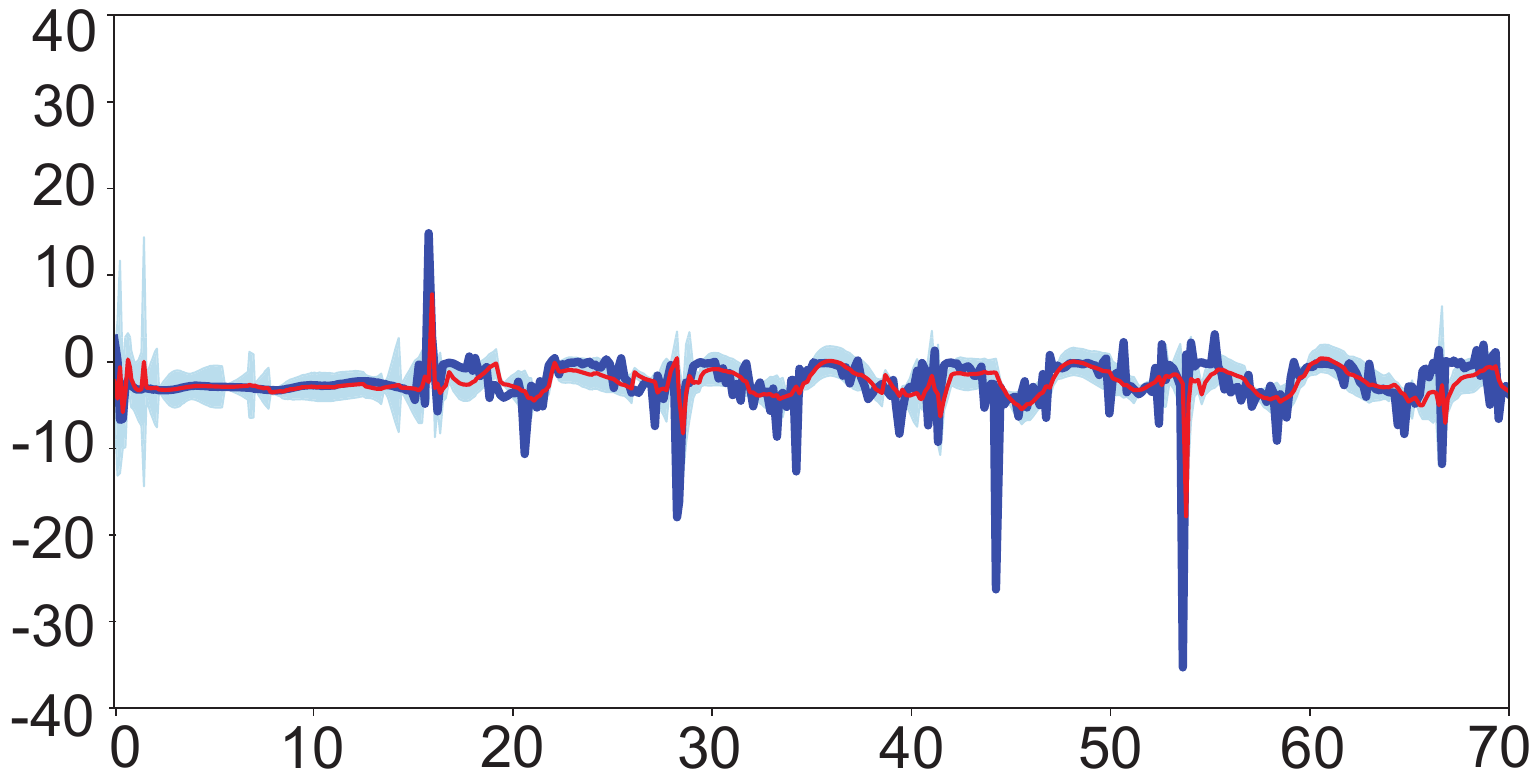}};
            \node at ($(fig68.south)+(0,-0.35)$) {\normalsize (e) dynamic obstacle with SOGP};
            \node[rectangle, rounded corners=1pt,minimum width=2.1cm,minimum height=0.3cm,draw=gray,fill=white!10](box) at ($(fig68.north east)+(-1.3,-0.38)$) {};
            \node (label) at ($(box.west)+(0.9,0.03)$) {\tiny $\Delta B^{(m)}$};
            \node (label2) at ($(box.east)+(-0.2,0.03)$) {\tiny $m^*$};
            \draw[draw=blue,line width=1pt] ($(label.west)+(-0.25,-0.05)$) -- ($(label.west)+(0.05,-0.05)$) node (legend){};
            \draw[draw=red,line width=1pt] ($(label2.west)+(-0.25,-0.05)$) -- ($(label2.west)+(0.05,-0.05)$);
            \draw[draw=orange!90!black,dashed,line width=1pt] ($(fig68.north)+(-1.18,-0.2)$) -- ($(fig68.south)+(-1.18,0.35)$);
            \node(text) at ($(fig68.south)+(0.15,0.55)$) {\footnotesize Obstacle approaching};
            \draw[arrow,red] (text.north) -- ++(-1.2,0.3);
            \node[rotate=90] at ($(fig68.west)+(0.,0)$) {\scriptsize TUC$[m\slash s^2]$};
            \node at ($(fig68.south)+(0,-0.02)$) {\scriptsize time$[s]$};
        \end{scope}

        \begin{scope}[xshift=2.95cm, yshift=-3cm]
            \node[above right] (fig67) at (0,0){\includegraphics[width=0.305\linewidth,height=0.147\linewidth]{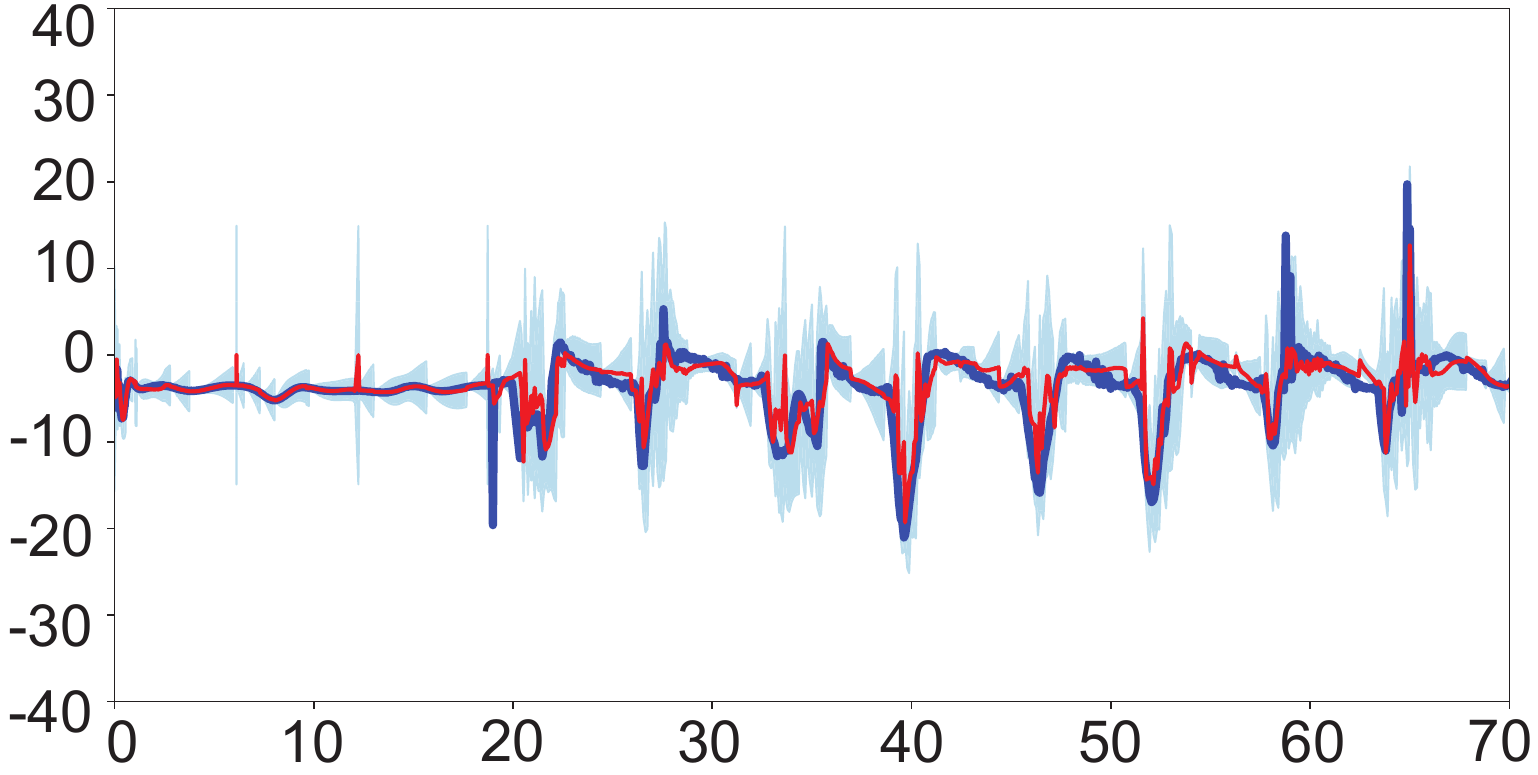}};
            \node at ($(fig67.south)+(0,-0.35)$) {\normalsize (f) dynamic obstacle with AFVSGP};
            \node[rectangle, rounded corners=1pt,minimum width=2.1cm,minimum height=0.3cm,draw=gray,fill=white!10](box) at ($(fig67.north east)+(-1.33,-0.35)$) {};
            \node (label) at ($(box.west)+(0.9,0.03)$) {\tiny $\Delta B^{(m)}$};
            \node (label2) at ($(box.east)+(-0.2,0.03)$) {\tiny $m^*$};
            \draw[draw=blue,line width=1pt] ($(label.west)+(-0.25,-0.05)$) -- ($(label.west)+(0.05,-0.05)$) node (legend){};
            \draw[draw=red,line width=1pt] ($(label2.west)+(-0.25,-0.05)$) -- ($(label2.west)+(0.05,-0.05)$);
            \draw[draw=orange!90!black,dashed,line width=1pt] ($(fig67.north)+(-0.98,-0.2)$) -- ($(fig67.south)+(-0.98,0.35)$);
            \node(text) at ($(fig67.south)+(0.3,0.55)$) {\footnotesize Obstacle approaching};
            \draw[arrow,red] (text.north) -- ++(-1.2,0.3);
            \node[rotate=90] at ($(fig67.west)+(0,0)$) {\scriptsize TUC$[m\slash s^2]$};
            \node at ($(fig67.south)+(0,-0.02)$) {\scriptsize time$[s]$};
        \end{scope}
    \end{tikzpicture}
  \caption{\quad Comparative analysis of predictive distributions across various learning methods under Gazebo simulation platform. Blue lines indicate the discrepancies between nominal estimates and actual model values, while red lines represent predictions from different models. A higher degree of alignment between red and blue lines suggests superior model performance. TUC represents the uncertain component with HOCBF and the blue shaded region represents the 95\% confidence interval (the same applies to other figures). }
  \label{fig2}
\end{figure*}
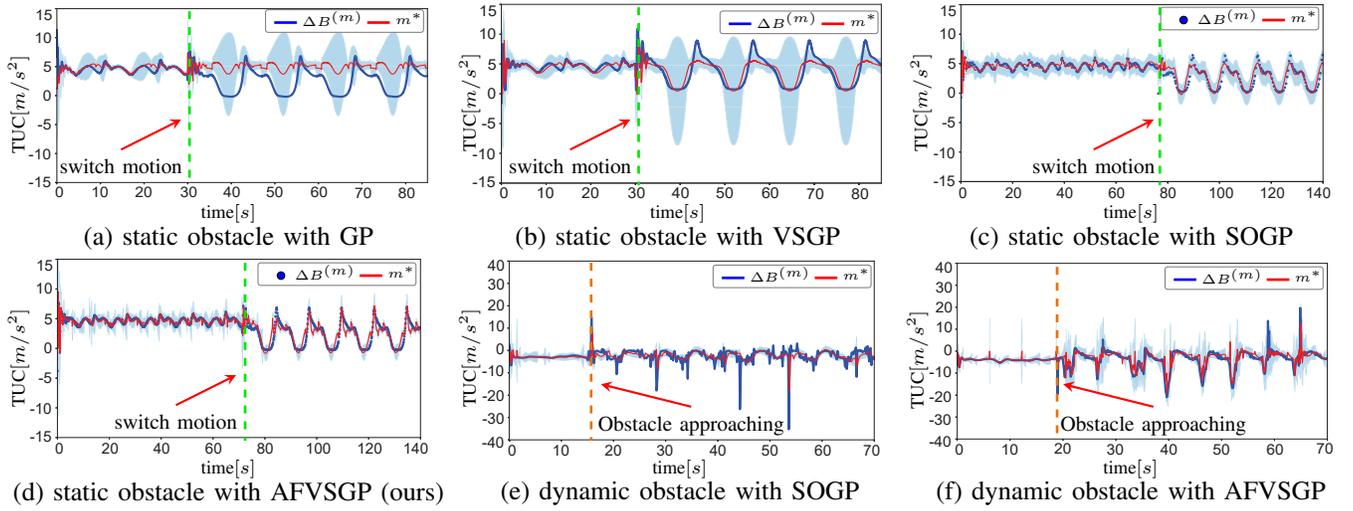
\subsection{Variational Sparse Gaussian Processes}
The VSGP approach outlined in \cite{titsias2009variational} employs inducing points as variational parameters to approximate the true GP posterior, thereby reducing overfitting and enabling the SGP to converge to its original form. Consider a data set $ ({\boldsymbol{\xi}}, {\boldsymbol{z}} ) $, which is derived from the relationship $z_{j} = f({\boldsymbol{\xi}}_j)+\varepsilon_n $ where $ f: \mathbb{R}^b \rightarrow \mathbb{R} $. The prior information is \( {\boldsymbol{\xi}} :=[{\boldsymbol{\xi}}_1, {\boldsymbol{\xi}}_2, \ldots, {\boldsymbol{\xi}}_N]^T \in \mathbb{R}^{N \times b} \), while \( {\boldsymbol{z}} \) is defined as \( {\boldsymbol{z}} := [z_1, z_2, \ldots, z_N]^T \in \mathbb{R}^N \). Here, \( N \) denotes the number of training data points, \( b \) specifies the input dimensionality, and the term \( \varepsilon_n \), which is distributed as \( \varepsilon_n \sim \mathcal{N}(0, \sigma_n^2) \), accounts for Gaussian white measurement noise. Next, we will delve into the distinct features of this algorithm. A collection of inducing inputs are denoted by \( \mathbf{O} = \left[\mathbf{o}_1, \ldots, \mathbf{o}_M\right] \), where each \( \mathbf{o}_i \) belongs to \( {\boldsymbol{\xi}} \). These inputs correspond to latent variables \( \mathbf{f}_o = \left[f\left(\mathbf{o}_1\right), \ldots, f\left(\mathbf{o}_M\right)\right]^{\top} \). Under the assumption that \( \mathbf{f}_o \) serves as a sufficient statistic for \( \mathbf{f} \), the predictive posterior of the VSGP, given by \eqref{my_equ12}, can be approximated using \eqref{my_equ13}, as detailed subsequently.
\begin{equation} \label{my_equ12}
p\left(f_* \mid {\boldsymbol{z}} \right)=\iint p\left(f_* \mid \mathbf{f}_o, \mathbf{f}\right) p\left(\mathbf{f} \mid \mathbf{f}_o, {\boldsymbol{z}} \right) p\left(\mathbf{f}_o \mid {\boldsymbol{z}} \right) d \mathbf{f} d \mathbf{f}_o\end{equation}
\begin{equation} \label{my_equ13}
\begin{aligned}
q\left(f_*\right)&=\int p\left(f_* \mid \mathbf{f}_o\right) q\left(\mathbf{f}_o\right) d \mathbf{f}_o
\end{aligned}
\end{equation}
where $p\left(f_* \mid \mathbf{f}_o, \mathbf{f}\right)$ is approximated by $p\left(f_* \mid \mathbf{f}_o\right)$, and posterior distribution $p\left(\mathbf{f}_o \mid {\boldsymbol{z}} \right)$ is approximated by a gaussian variational distribution $q\left(\mathbf{f}_o\right)$ with mean ${\boldsymbol{\mu}}$ and covariance ${\boldsymbol{A}}$, and the approximate predictive posterior is shown as
\begin{equation} \label{my_equ14}
\begin{aligned}
q\left(f_*\right) & =\mathcal{N}\left(f_* \mid m_*, v_*\right) \\
m_* & =\mathbf{k}_{* o} \mathbf{K}_{o o}^{-1} \boldsymbol{\mu} \\
v_* & =k_{* *}-\mathbf{k}_{* o} \mathbf{K}_{o o}^{-1} \mathbf{k}_{o *}+\mathbf{k}_{* o} \mathbf{K}_{o o}^{-1} \mathbf{A} \mathbf{K}_{o o}^{-1} \mathbf{k}_{o *},
\end{aligned}
\end{equation}
where $\mathbf{k}_{* o}=\mathbf{k}_{o *}^{\top}$ is a row vector with the kernel products between $ \xi^*$ and $\mathbf{O}$, and covariance $\mathbf{K}_{o o}$ is the matrix that contain all kernel dot products $k (\mathbf{o}, \mathbf{o}^{\prime})$ for inducing inputs, and $k_{* *} = k({\boldsymbol{\xi}}_{*}, {\boldsymbol{\xi}}_{*})$ is the kernel product of a query point.

To obtain the variational parameters  ${\boldsymbol{\mu}}$ and ${\boldsymbol{A}}$, authors in \cite{titsias2009variational} firstly introduced the subsequent variational lower bound for the authentic log marginal likelihood.
\begin{equation} \label{my_equ15}
\log p(\mathbf{z}) \geq F_V\left(\mathbf{O}, q\left(\mathbf{f}_o\right)\right)=\int q\left(\mathbf{f}_o\right) \log \frac{G\left(\mathbf{f}_o, \mathbf{z}\right) p\left(\mathbf{f}_o\right)}{q\left(\mathbf{f}_o\right)} d \mathbf{f}_o
\end{equation}
where the link $\log G\left(\mathbf{f}_o, \mathbf{z}\right) =\log \mathcal{N}\left(\mathbf{z} \mid \mathbf{K}_{\xi o} \mathbf{K}_{o o}^{-1} \mathbf{f}_o, \sigma^2 \mathbf{I}\right) -(1/2 \sigma^2) \operatorname{tr}\{\mathbf{K}_{\xi \xi}-\mathbf{K}_{\xi o} \mathbf{K}_{o o}^{-1} \mathbf{K}_{o \xi}\}$ is shown with the trance operator $\operatorname{tr}\{\cdot\}$. Subsequently, authors differentiate \eqref{my_equ15} with respect to $q\left(\mathbf{f}_o\right)$ and equate it to zero to obtain the optimal expression shown as follows.
\begin{equation} \label{my_equ17}
\begin{aligned}
q\left(\mathbf{f}_o\right) & =\mathcal{N}\left(\mathbf{f}_o \mid \boldsymbol{\mu}, \mathbf{A}\right)
\end{aligned}
\end{equation}
where $\boldsymbol{\mu} =\sigma^{-2} \mathbf{K}_{o o} \mathbf{B K}_{o \xi} \mathbf{z}$, $\mathbf{A} =\mathbf{K}_{o o} \mathbf{B K}_{o o} $, $\mathbf{B}=\left(\mathbf{K}_{o o}+\sigma^{-2} \mathbf{K}_{o \xi} \mathbf{K}_{\xi o}\right)^{-1}$. To further refine the hyperparameters, including the noise variance and the kernel parameters, and to update the set of $\mathbf{O}$, the methodology outlined in \cite{titsias2009variational} utilized Jensen's inequality. This was employed to eliminate the distribution $q\left(\mathbf{f}_o\right)$, resulting in a collapsed version of the bound presented in \eqref{my_equ15}.
\begin{equation} \label{my_equ18}
\begin{aligned}
F_V(\mathbf{O}) & =\log \mathcal{N}\left(\mathbf{z} \mid \mathbf{0}, \sigma^2 \mathbf{I}+\mathbf{K}_{\xi o} \mathbf{K}_{o o}^{-1} \mathbf{K}_{o \xi}\right) \\
& -\frac{1}{2 \sigma^2} \operatorname{tr}\left\{\mathbf{K}_{\xi \xi}-\mathbf{K}_{\xi o} \mathbf{K}_{o o}^{-1} \mathbf{K}_{o \xi}\right\}
\end{aligned}
\end{equation}

Ultimately, the optimal values are obtained by maximizing the bound in \eqref{my_equ18}, and both the computational and optimization burden for  \eqref{my_equ17} and \eqref{my_equ18} are $\mathcal{O}\left(N M^2\right)$, respectively.

\section{Adaptive Online VSGP with the Affine Dot Product Kernel}
This section comprises three key components for our proposed novel adaptive fast VSGP (AFVSGP) framework: 1) offline hyperparameter training, 2) adaptive modification and forgetting of training data and inducing points, and 3) real-time model inference updates.
\subsection{Offline Training for Hyperparameter}
We begin by selecting a small subset of preliminary data, denoted by $P$ for offline hyperparameter training, where $P \ll N$ and $N$ is the typical training data size in standard GPs. Following this, we integrate a specialized kernel function \cite{castaneda2021gaussian} into the VSGP algorithm.
\begin{definition}
Affine Dot Product compound kernel: Define $k_a: \overline{\mathcal{X}} \times \overline{\mathcal{X}} \rightarrow \mathbb{R}$ given by
\begin{equation} \label{my_equ19}
k_a\left(\begin{bmatrix}
\xi \\
\bar{u}
\end{bmatrix},
\begin{bmatrix}
\xi^{\prime} \\
\bar{u}^{\prime}
\end{bmatrix}\right) = \bar{u}^T \operatorname{Diag}\left(k_1\left(\xi, \xi^{\prime}\right), \cdots, k_{m+1}\left(\xi, \xi^{\prime}\right)\right) \bar{u}^{\prime}
\end{equation}
as the Affine Dot Product (ADP) compound kernel of $(m+1)$ individual kernals $k_1, \ldots, k_{m+1}: \mathcal{X} \times \mathcal{X} \rightarrow \mathbb{R}$.
\end{definition}
\begin{proposition}
  Let $k_1, \ldots, k_{m+1}$ be a set of positive definite kernels. Then the ADP compound kernel $k_a$ inherits positive definiteness. Additionally, if $k_1, \ldots, k_{m+1}$ are bounded, $k_a$ remains bounded.
\end{proposition}

Building upon Thm. 1 in \cite{castaneda2021gaussian}, $\mathcal{H}_{k_a}(\mathcal{\xi} \times \mathcal{\bar{U}})$ is an Reproducing Kernel Hilbert Space (RKHS) whose reproducing
kernel is $k_a$ in Definition 4.  We include a forgetting factor $\phi$ into this term and start considering time instant $t$, so only data from initial sample at $t^{\prime}=1$ to $t$ are available, then, with the ADP kernal, \eqref{my_equ14} is rewritten as
\begin{equation} \label{my_equ20}
\begin{aligned}
q\left(f_*\right) & =\mathcal{N}\left(f_* \mid m_{*,\phi}, v_{*, \phi}\right) \\
m_{*,\phi} &= \sigma^{-2} \mathbf{z} ^{\top} \mathbf{\Lambda} \mathbf{K}_{\xi o} \mathbf{B}_{\phi} (\mathbf{k}_{*\bar{U}_{o}}^{\top}) \bar{u}_{*}\\
v_{*,\phi} & =\bar{u}_{*}^{\top}(k_{**}+ \mathbf{k}_{* \bar{U}_{o}} (\mathbf{B}_{\phi} - \mathbf{K}_{o o}^{-1}) \mathbf{k}_{\bar{U}_{o} *})\bar{u}_{*}
\end{aligned}
\end{equation}
where $\mathbf{z}_{j} = \mathcal{F}\left(\xi_j, \bar{u}_j\right)+\epsilon_n$, $\mathbf{K}_{\xi o} \in \mathbb{R}^{P \times M}$ is the gram matrix of $\bar{u}_{\xi}^T \operatorname{Diag}\left(\left[k_1\left(\xi, \xi^{\prime}_{o}\right), \cdots, k_{m+1}\left(\xi, \xi^{\prime}_{o}\right)\right]\right) \bar{u}^{\prime}_{o}$, $\mathbf{B}_{\phi}=\left(\mathbf{K}_{o o}+\sigma^{-2} \mathbf{K}_{ o \xi} \mathbf{K}_{\xi o}\right)^{-1} \in \mathbb{R}^{M \times M}$ and $\mathbf{K}_{o o}$ are also gram matrix of $\bar{u}_{o}^T \operatorname{Diag}\left(\left[k_1\left(\xi_{o}, \xi^{\prime}_{o}\right), \cdots, k_{m+1}\left(\xi_{o}, \xi^{\prime}_{o}\right)\right]\right) \bar{u}^{\prime}_{o}$, $\mathbf{z} \in \mathbb{R}^{P} $, and $(\mathbf{k}_{*\bar{U}_{o}}) \in \mathbb{R}^{(m+1) \times M}$ is given by
\begin{equation} \label{my_equ21}
\mathbf{k}_{*\bar{U}_{o}}=\left[\begin{array}{c}
K_{1 *} \
K_{2 *} \
\cdots \
K_{(m+1) *}
\end{array}\right]^{\top} \circ *\bar{U}_{o}
\end{equation}
\begin{equation} \label{my_equ22}
K_{i *}=\left[k_i\left(\xi_*, \xi_1\right), \cdots, k_i\left(\xi_*, \xi_M\right)\right]
\end{equation}
and $k_{**} = \text {Diag}\{k_{1}(\xi_{*}, \xi_{*}), \ldots, k_{m+1}(\xi_{*}, \xi_{*})\}$, and where $\boldsymbol{\Lambda}$ is a $P \times P$ diagonal matrix with $\mathbf{\Lambda}_{t^{\prime} t^{\prime}}=\phi^{P-t^{\prime}}$. To enhance clarity and promote an effective understanding of the text, we opt to utilize the symbol $k(\cdot)$ in lieu of $k_{a}(\cdot)$ in \eqref{my_equ19}. To update hyperparameters, equation \eqref{my_equ18} is rewritten as.

\begin{equation} \label{my_equ23}
\begin{aligned}
F_V^\phi(\mathbf{O}) & \propto \log \mathcal{N}\left(\mathbf{z} \mid \mathbf{0}, \mathbf{K}_{\xi o} \mathbf{K}_{o o}^{-1} \mathbf{K}_{o \xi} + \sigma^2 \mathbf{\Lambda}^{-1}\right)  \\
& -\frac{1}{2 \sigma^2} \sum_{t^{\prime}=1}^t\left(\phi^{t-t^{\prime}}-1\right) \log \left(2 \pi \sigma^2\right) \\
& +\frac{1}{2 \sigma^2} \sum_{t^{\prime}=1}^t \phi^{t-t^{\prime}}\left(k_{t^{\prime} t^{\prime}}-\mathbf{k}_{t^{\prime} o} \mathbf{K}_{o o}^{-1} \mathbf{k}_{o t^{\prime}}\right)
\end{aligned}
\end{equation}
where $\mathbf{z}=\left[z_1, \ldots, z_t\right]^{\top}$ and each kernel is replaced by the ADP compound kernel. Employing \eqref{my_equ20} and \eqref{my_equ23}, we are capable of learning from the collected $P$ data offline, simultaneously acquiring the hyperparameters and the approximate predictive posterior. This offline computational process carries a computational burden of $\mathcal{O}\left(P M^2\right)$.
\begin{algorithm}
\caption{AFVSGP-HOCBF-SOCP}
\small{
\begin{algorithmic}[1]
\State \textbf{Inputs:} New sample $\left((\mathbf{\xi}_{t+1}, {\bar{u}}_{t+1}), z_{t+1}\right)$, offline model $q\left(f_*\right)$ with \eqref{my_equ20}, window $P$, thresholds $R_{th}$, $R_{thm}$.
\State Update training set and compute $(m_{*,\phi})_{t+1} $and  $(v_{*,\phi})_{t+1}$ with \eqref{my_equ24} for new data sample.
\If{$P_{th} > \epsilon$ [see \eqref{my_equ27}]} \State set the new sample in inducing points as $\mathbf{o}_{k+1}$ and update the predictive distribution with \eqref{my_equ29}. \EndIf
\If{$\mathbf{o}_{k-1}$ with less information [see \eqref{my_equ28}]} \State Update the predictive distribution with \eqref{my_equ32} and get the new model $q\left(f_*\right)$. \EndIf
\State Compute the safety-aware controller with \eqref{my_equ05} and get new sample with \eqref{my_equ48}.
\State \textbf{return} $\text{Safety control input}$ $\bar{u}_{*}$.
\end{algorithmic}}
\end{algorithm}
\begin{table}[h]
\caption{\quad Comparison of different methods while avoiding a static obstacle under Gazebo simulation platform. }
\label{table:comparison}
\centering
\begin{tabular}{l|c|c|c|c}
\hline
Model & AFVSGP & VSGP & GP & SOGP \\
\hline
MSE & \textbf{0.5863} & 1.4844 & 6.6029 & 0.8118 \\
\hline
Tr. Time(s) & \textbf{0.00613} & 0.03562 & / & 0.08724 \\
\hline
\end{tabular}
\end{table}
\subsection{Adaptive Online Learning for Data and Inducing Points}
Following offline analysis and hyperparameter optimization, we transition to online adaptive updates. This is essential for real-time posterior approximation upon the arrival of each new training sample \((\mathbf{\xi}_{t+1}, {\bar{u}}_{t+1}), z_{t+1})\). The oldest sample is replaced to maintain a fixed-size training set \(P\), and both the mean and variance are updated at \(t+1\).
\begin{equation} \label{my_equ24}
\begin{aligned}
&(m_{*,\phi})_{t+1} = \sigma^{-2} (\mathbf{z} ^{\top} \mathbf{\Lambda} \mathbf{K}_{\xi o})_{t+1} (\mathbf{B}_{\phi})_{t+1} (\mathbf{k}_{*\bar{U}_{o}}^{\top}) \bar{u}_{*}\\
&(v_{*,\phi})_{t+1}  =\bar{u}_{*}^{\top}(k_{**}+ \mathbf{k}_{* \bar{U}_{o}} ((\mathbf{B}_{\phi})_{t+1} - \mathbf{K}_{o o}^{-1}) \mathbf{k}_{\bar{U}_{o} *})\bar{u}_{*}
\end{aligned}
\end{equation}
where
\begin{equation} \label{my_equ25}
\begin{aligned}
&(\mathbf{B}_{\phi})_{t+1}=((1-\phi) \mathbf{K}_{oo}+\phi(\mathbf{B}_{\phi})_{t}^{-1}+\sigma^{-2} \mathbf{k}_{t+1, o}^{\top} \mathbf{k}_{t+1, o}\\
&\qquad \quad \quad \;\;-\phi^{\top}\mathbf{k}_{t-T, o}^{\top} \mathbf{k}_{t-T, o})^{-1} \nnum
\end{aligned}
\end{equation}
\begin{equation} \label{my_equ26}
\begin{aligned}
&(\mathbf{z} ^{\top} \mathbf{\Lambda} \mathbf{K}_{\xi o})_{t+1} = \phi(\mathbf{z} ^{\top} \mathbf{\Lambda} \mathbf{K}_{\xi o})_{t} + z_{t+1} \mathbf{k}_{t+1, o}- z_{t-T}\phi \mathbf{k}_{t-T, o}\nnum
\end{aligned}
\end{equation}

If the previous iteration $\mathbf{B}_{\phi}$ and $\mathbf{z} ^{\top} \mathbf{\Lambda} \mathbf{K}_{\xi o}$ are saved, the complexity of operations $(\mathbf{B}_{\phi})_{t+1}$ and $(\mathbf{z} ^{\top} \mathbf{\Lambda} \mathbf{K}_{\xi o})_{t+1}$ are $O\left(M^3\right)$ and $O(M)$, respectively. After revising the training set, we use established metrics to select the most representative inducing points, as inspired by \cite{gomez2023adaptive,csato2002sparse,9961103}. These points are added to the inducing set, while the least relevant ones are removed. The value of the top representative inducing point should exceed the subsequent regularization term linked to the adaptive collapsed marginal likelihood, detailed in \eqref{my_equ23}.
\begin{equation} \label{my_equ27}
P_{th} =\sum_{t^{\prime}=1}^t \phi^{t-t^{\prime}} (k_{t^{\prime} t^{\prime}}-\mathbf{k}_{t^{\prime} o} \mathbf{K}_{oo}^{-1} \mathbf{k}_{o t^{\prime}})
\end{equation}

Thus, if a newly introduced training point results in a value from \eqref{my_equ27} exceeding a user-defined threshold $\epsilon$, i.e., $P_{th} > \epsilon$, it is added to the inducing set. Additionally, we use the same equation to assess the representativeness of existing points. The least representative point is identified and removed based on criteria for each $P_{thm}, m=1, \ldots, M $.
\begin{equation} \label{my_equ28}
P_{thm}=\sum_{t^{\prime}=1}^t \phi^{t-t^{\prime}} \mathbf{k}_{ t^{\prime} m} \mathbf{K}_{m m}^{-1} \mathbf{k}_{m t^{\prime}}
\end{equation}
where $P_{\mathrm{th}}=\sum_{t^{\prime}=1}^t \phi^{t-t^{\prime}} k_{t^{\prime} t^{\prime}}-\sum_{m=1}^M P_{thm}$. On incorporating the new sample $\left((\mathbf{\xi}_{k+1}, {\bar{u}}_{k+1}), z_{k+1}\right)$ as $\mathbf{o}_{k+1}$ into the inducing inputs $\mathbf{O}=\left[\mathbf{o}_1, \ldots, \mathbf{o}_M\right]$, the approximate predictive posterior is updated using the following procedure.
\begin{equation} \label{my_equ29}
\begin{aligned}
(m_{*,\phi})_{k+1} &= \sigma^{-2} \mathbf{z} ^{\top} \mathbf{\Lambda} (\mathbf{K}_{\xi o})_{k+1} (\mathbf{B}_{\phi})_{k+1} (\mathbf{k}_{*\bar{U}_{o}}^{\top})_{k+1} \bar{u}_{*}\\
(v_{*,\phi})_{k+1} & =\bar{u}_{*}^{\top}(k_{**}+ (\mathbf{k}_{* \bar{U}_{o}})_{k+1} ((\mathbf{B}_{\phi})_{k+1} \\
&\quad - (\mathbf{K}_{o o}^{-1})_{k+1}) (\mathbf{k}_{\bar{U}_{o} *})_{k+1})\bar{u}_{*} \text{,}
\end{aligned}
\end{equation}
where $(\mathbf{k}_{*\bar{U}_{o}}^{\top})_{k+1}$ is straightforward since it only implies adding a row and new element to this vector and matrix, respectively. The update of $(\mathbf{B}_{\phi})_{k+1}$ can be effectively computed from $(\mathbf{B}_{\phi})_{k}$ using the properties of the block matrix inversion.
\begin{equation} \label{my_equ30}
\begin{aligned}
\left(\mathbf{B}_{\phi}\right)_{k+1} & =\left(\begin{array}{cc}
(\mathbf{B}_{\phi}^{-1})_k & (\mathbf{b}_{\phi})_{k+1} \\
(\mathbf{b}_{\phi}^{\top})_{k+1} & (b_{\phi})_{k+1}
\end{array}\right)^{-1}
\end{aligned}
\end{equation}
where $ \left(b_{\phi}\right)_{k+1}=k_{k+1, k+1}+\sigma^{-2} \mathbf{k}_{k+1, \xi} \boldsymbol{\Lambda} \mathbf{k}_{\xi, k+1}$ and  $\left(\mathbf{b}_{\phi}\right)_{k+1} =\mathbf{k}_{o, k+1}+\sigma^{-2}\left(\mathbf{K}_{o \xi}\right)_k \boldsymbol{\Lambda} \mathbf{k}_{\xi, k+1}$. Consequently, employing \eqref{my_equ29} and \eqref{my_equ30}, the time complexity associated with the addition of a new inducing point is found to be $\mathcal{O}\left(M^3\right)$ due to compute the inversions of $\left(\mathbf{B}_\phi\right)_k$ and $\left(\mathbf{K}_{o o}\right)_k^{-1}$. In addition, when we delete a inducting point in the set, the mean and variance is rewritten as
\begin{equation} \label{my_equ32}
\begin{aligned}
(m_{*,\phi})_{k-1} &= \sigma^{-2} \mathbf{z} ^{\top} \mathbf{\Lambda} (\mathbf{K}_{\xi o})_{k-1} (\mathbf{B}_{\phi})_{k-1} (\mathbf{k}_{*\bar{U}_{o}}^{\top})_{k-1} \bar{u}_{*}\\
(v_{*,\phi})_{k-1} & =\bar{u}_{*}^{\top}(k_{**}+ (\mathbf{k}_{* \bar{U}_{o}})_{k-1} ((\mathbf{B}_{\phi})_{k-1} \\
&\quad - (\mathbf{K}_{o o}^{-1})_{k-1}) (\mathbf{k}_{\bar{U}_{o} *})_{k-1})\bar{u}_{*}
\end{aligned}
\end{equation}
where $(\mathbf{B}_{\phi})_{k-1}=\left((\mathbf{K}_{oo})_{k-1}+\sigma^{-2} (\mathbf{K}_{o \xi})_{k-1} \Lambda (\mathbf{K}_{\xi o})_{k-1}\right)^{-1}$. Other $(k-1)$ elements are delating a row, a column, or an element to its
vector and matrix, so we omit its process.

\section{Proposed Safety Controller}
In this section, we aim to integrate this AFVSGP framework into \eqref{my_equ04} to enhance safety. The $m$th derivatives of the HOCBF for the affine system can be expressed as
\begin{equation} \label{my_equ44}
\begin{aligned}
B^{(m)}(\xi, \bar{u}) =& \mathcal{F}(\xi) \bar{u} + S(b(\xi)) + \alpha_m(\psi_{m-1}(\xi))
\end{aligned}
\end{equation}
where $\mathcal{F}(\xi) = [L_f^m b(\xi) \; \; L_g L_f^{m-1} b(\xi)]$ and $\bar{u} =[1 \; \; u^{\top}]^{\top}$.

The nominal estimates of their values are given by
\begin{equation} \label{my_equ45}
\begin{aligned}
B^{(m)}_{0}(\xi, \bar{u}) =& \mathcal{F}_{0}(\xi) \bar{u} + S(b(\xi)) + \alpha_m(\psi_{m-1}(\xi))
\end{aligned}
\end{equation}
where $\mathcal{F}(\xi) = [L_{f0}^m b(\xi) \; \; L_{g0} L_{f0}^{m-1} b(x) ]$.

We introduce the function $\Delta{B^{(m)}}: \mathcal{X} \times \mathbb{R}^m \rightarrow \mathbb{R}$ to represent the discrepancy between \eqref{my_equ44} and \eqref{my_equ45}.
\begin{equation} \label{my_equ46}
\begin{aligned}
\Delta B^{(m)}(\xi, \bar{u}) & :=B^{(m)}(\xi, \bar{u})-B^{(m)}_0(\xi, \bar{u}).
\end{aligned}
\end{equation}
Then, the HOCBF constraints become
\begin{equation} \label{my_equ47}
\begin{aligned}
B^{(m)}_{0}(\xi, \bar{u}) + \Delta B^{(m)}(\xi, \bar{u}) \geq 0
\end{aligned}
\end{equation}

It should be noted that the uncertainties, $\Delta B^{(m)}(\xi, \bar{u})$, are scalar values. This results in a dimensionally reduced learning problem compared to directly learning the dynamics, $\mathcal{F}(\xi)$. To approximate $\Delta B^{(m)}(\xi, \bar{u})$, we can collect trajectories from the actual plant. These measurements are provided as follows.
\begin{equation} \label{my_equ48}
\begin{aligned}
z_j^{\Delta B^{(m)}} = & (B^{(m-1)}(\xi(t+\Delta t)) - B^{(m-1)}(\xi(t)))/\Delta t \\
                      & - {B_{0}^{(m)}}\left(\xi_j, u_j\right)
\end{aligned}
\end{equation}
where $\xi_j=(\xi(t+\Delta t)+\xi(t)) / 2$ represents the state's average over the interval $(t, t+\Delta t)$. The term $u_j$ denotes the control input within the same timeframe. The parameter $\Delta t$ defines the sampling interval. Meanwhile, $z_j^{\Delta B^{(m)}}$ serves as an approximation of the measurements for $\Delta B^{(m)}(\xi, \bar{u})$ for $j=1, \ldots, N$. Given these datasets, the regression challenges concerning $\Delta B^{(m)}(\xi, \bar{u})$ can be effectively framed as supervised learning tasks.
\begin{assumption}
  The candidate function $\Delta B^{(m)}(\xi, \bar{u})$ belong to the RKHS \cite{wendland2004scattered}, associated to the reproducing kernel function \eqref{my_equ19}. Therefore, it has bounded RKHS norm
  with respect to $k_{a}$, i.e., $\|\Delta B^{(m)}(\xi, \bar{u})\|_{ka}^2  \leq \|\Delta\|_{ka}$.
\end{assumption}

According to Thm. 2 in \cite{castaneda2021gaussian}, each subsequent inequality is satisfied with a probability of at least $1-\delta$ for all $\xi_i \in \xi$ and $ \bar{u} \in \mathcal{\bar{U}}$.
\begin{equation} \label{my_equ49}
\begin{aligned}
& B^{(m)}(\xi, \bar{u}) \geq B^{(m)}_{0}(\xi, \bar{u})+\mu_{\Delta B}(\xi, \bar{u})-\beta \sigma_{\Delta B}(\xi, \bar{u})
\end{aligned}
\end{equation}
where $\beta:=\left(2 \|\Delta\|_{ka}^2+300 \kappa_{P+1} \ln ^3((P+1) / \delta)\right)^{0.5}$, with $P$ is the total number of data, $\delta \in(0,1)$, $\kappa_{P+1}$ represents the maximum information for which a uniform bound can be established, as detailed in \cite{srinivas2012information}, after obtaining $P+1$ data points. $\mu_{\Delta B}$ and $\sigma_{\Delta B}$ are the mean and standard deviation of the GP prediction of $ \Delta B^{(m)}(\xi_{*}, \bar{u}_{*})$, given by \eqref{my_equ32}.

We provide a detailed expression for $\mu_{\Delta B}$ and $\sigma_{\Delta B}$ as follows.
\begin{equation} \label{my_equ51}
\begin{aligned}
& \mu_{\Delta B}\left(\xi, u \mid \mathbb{D}_N\right)=b_{\Delta B}^{\top}\left(\xi \mid \mathbb{D}_N\right)\left[1 \; \; u^{\top}\right]^{\top} \\
& \sigma_{\Delta B}\left(\xi, u \mid \mathbb{D}_N\right)= \left[1\; \;  u^{\top}\right]\Sigma_{\Delta B}\left(\xi \mid \mathbb{D}_N\right)\left[1\; \; u^{\top}\right]^{\top}
\end{aligned}
\end{equation}
where $b_{\Delta B}^{\top} = \sigma^{-2} \mathbf{z} ^{\top} \mathbf{\Lambda} (\mathbf{K}_{\xi o})_{k-1} (\mathbf{B}_{\phi})_{k-1} (\mathbf{k}_{*\bar{U}_{o}}^{\top})_{k-1}$ and $\Sigma_{\Delta B} = k_{**}+ (\mathbf{k}_{* \bar{U}_{o}})_{k-1} ((\mathbf{B}_{\phi})_{k-1}
- (\mathbf{K}_{o o}^{-1})_{k-1})$ $(\mathbf{k}_{\bar{U}_{o} *})_{k-1}$. The uncertainty-aware control for safety of system \eqref{my_equ1} is finally transformed as an AFVSGP-HOCBF-SOCP problem shown as follows, and detailed in Alg. 1 and Fig. \ref{fig1}.
\begin{equation} \label{my_equ05}
\begin{aligned}
u(x)=& \underset{u \in \mathbb{R}^m}{\operatorname{argmin}}\;  \frac{1}{2}\|\bar{u}-\bar{u}_{\rm nom}\|^2  \\
\text { s.t. } & \mathcal{F}_{0}(\xi) \bar{u} + S(b(\xi)) + \alpha_m(\psi_{m-1}(\xi))+b_{\Delta B}^{\top} \bar{u}\\
& \qquad \qquad \qquad - \beta \bar{u}^{\top} \Sigma_{\Delta B} \bar{u} \geq 0
\end{aligned}
\end{equation}

Equation \eqref{my_equ05} adheres to the characteristics of a Second-Order Cone Program (SOCP). Numerous studies have emphasized its local Lipschitz continuity, underscoring the convex nature of the problem in data-driven safety-critical control when implementing GP-CBFs \cite{dean2021guaranteeing,buch2021robust,castaneda2021gaussian}. Moreover, once optimal hyperparameters have been determined through offline learning, our proposed framework guarantees an online learning time complexity of \(O\left(M^3\right)\), rendering the method highly applicable for real-time scenarios.
\section{Simulation and Real-World Experiments}
\begin{figure}[!t]
  \centering
  \begin{tikzpicture}
      \begin{scope}[xshift=0.05cm]
          \node[above right] (fig422) at (-2.2,0){\includegraphics[height=0.28\linewidth]{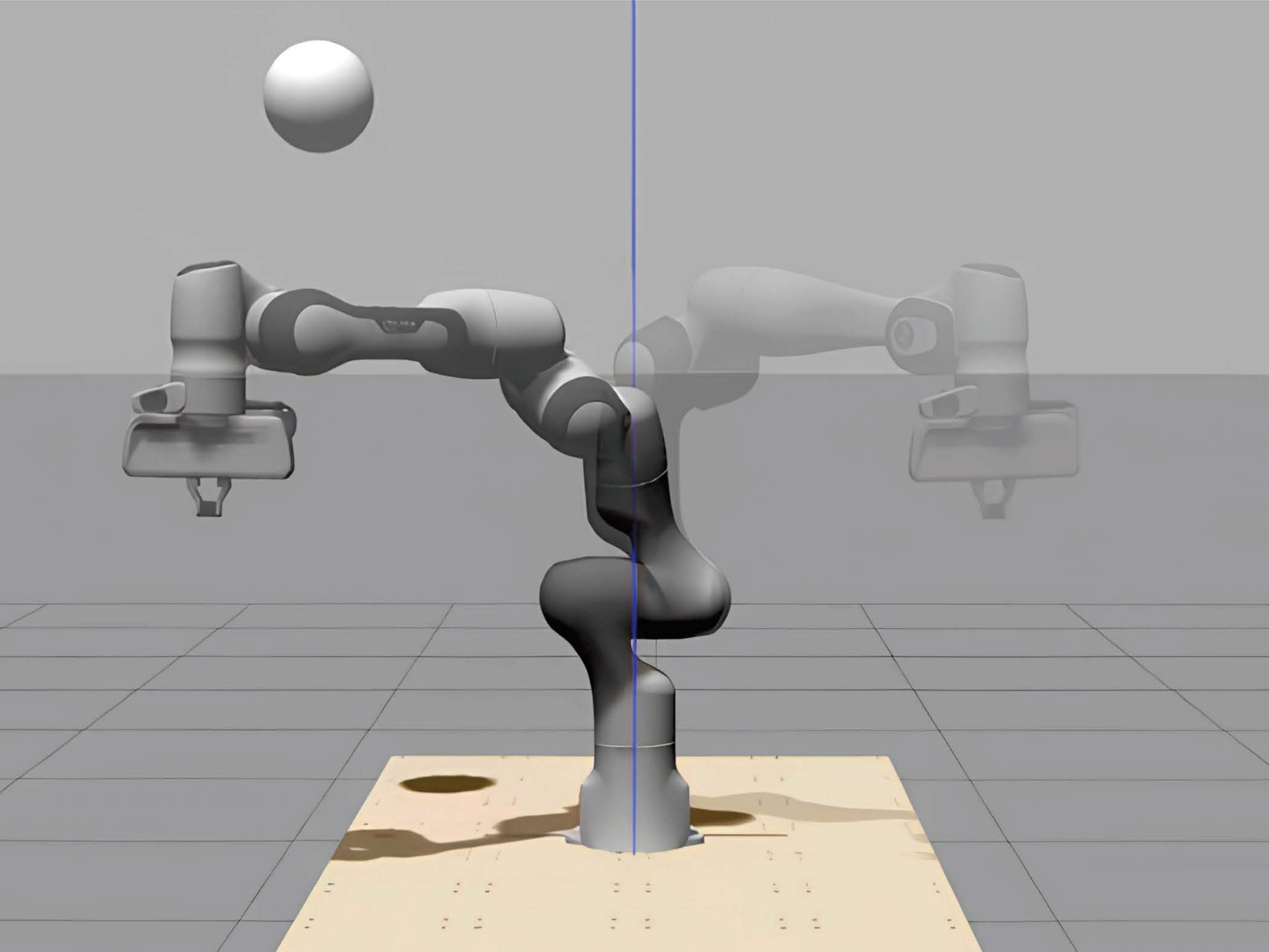}};
        \node at ($(fig422.south)+(0,-0.1)$) {\normalsize (a) along the y-axis};
        \draw[stealth-stealth, red, line width=0.8pt] ($(fig422.west)+(0.75,-0.1)$) -- ($(fig422.east)+(-0.9,-0.1)$);
        \draw[fill=green,draw=green] ($(fig422.west)+(0.7,-0.1)$) circle (1.5pt);
        \draw[fill=green,draw=green] ($(fig422.east)+(-0.85,-0.1)$) circle (1.5pt);
        \node[red] at ($(fig422.center)+(-0.1,0.1)$) {\scriptsize Train model};
        \node[yellow] at ($(fig422.center)+(-0.1,-0.25)$) {\scriptsize Access model};
      \end{scope}

    \begin{scope}[xshift=0.4cm]
        \node[above right] (fig433) at (2.2,0){\includegraphics[height=0.28\linewidth]{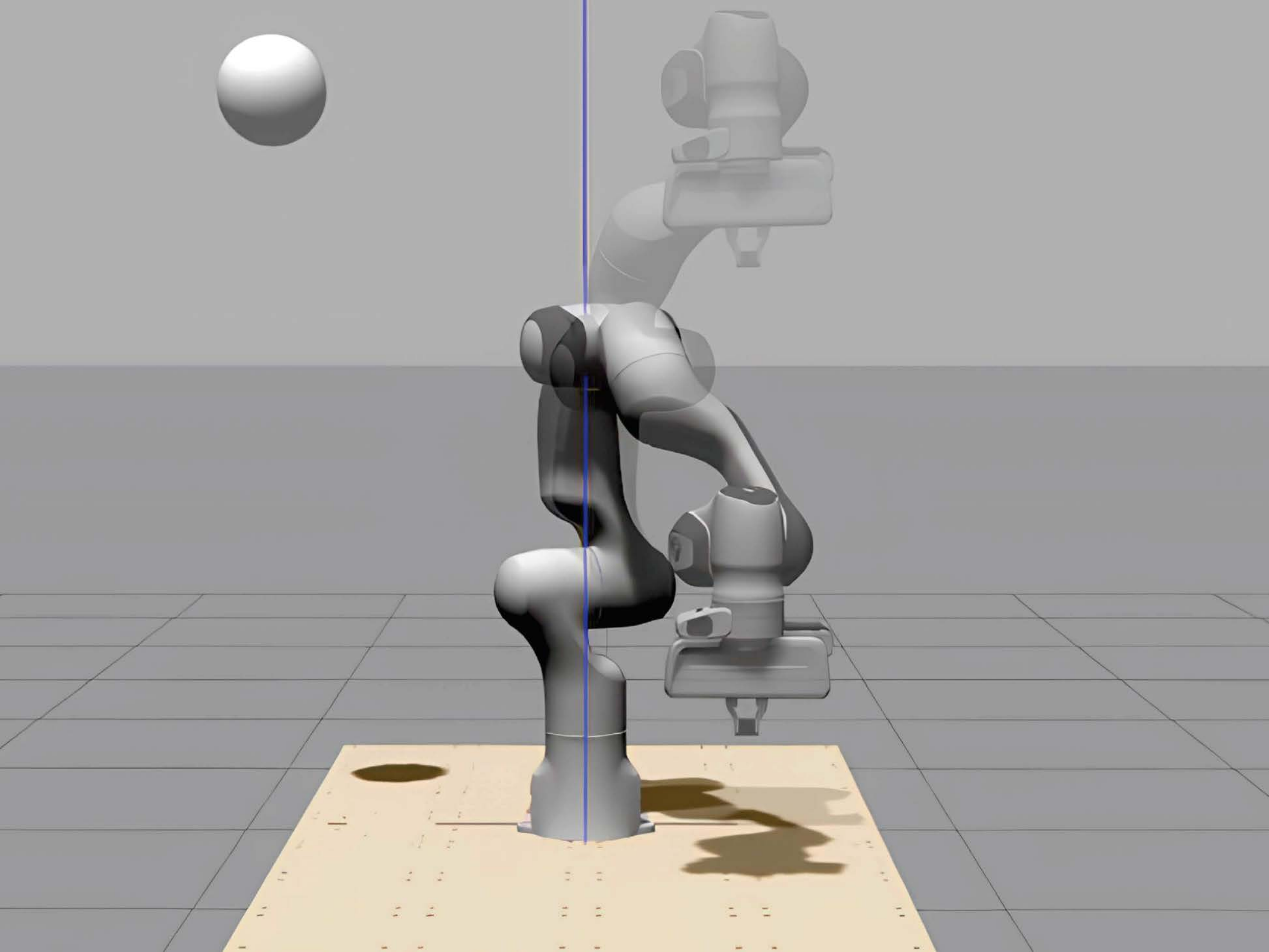}};
    \node at ($(fig433.south)+(0,-0.1)$) {\normalsize (b) along the z-axis};
    \draw[stealth-stealth, red, line width=0.8pt] ($(fig433.east)+(-1.47,0.5)$) -- ($(fig433.east)+(-1.47,-0.5)$);
    \draw[fill=green,draw=green] ($(fig433.east)+(-1.47,0.6)$) circle (1.5pt);
    \draw[fill=green,draw=green] ($(fig433.east)+(-1.47,-0.6)$) circle (1.5pt);
    \node[yellow, align=center] at ($(fig433.east)+(-0.8,-0.)$) {\scriptsize Unseen data};
    \node[minimum width=1.5cm,minimum height=0.45cm,draw=white, fill=white](sbox) at ($(fig433.west)+(-0.64,0.12)$) {};
    \draw[-stealth,cyan, line width=0.8pt] ($(sbox.west)+(0.1,-0.12)$) -- ($(sbox.east)+(-0.1,-0.12)$) node[midway, above]{\scriptsize switch motion};
    \end{scope}
  \end{tikzpicture}
  \caption{\quad Methods are initially trained and assessed using the end-effector's data on the y-axis. Subsequently, the motion is shifted to the z-axis to evaluate adaptability with previously unseen data.}
  \label{fig3}
  \vspace{-1.2em}
\end{figure}
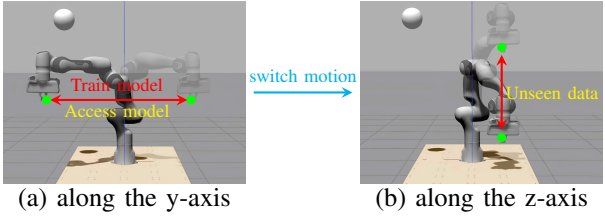

In this section, we evaluate obstacle avoidance tasks for a Franka robot in both Gazebo platform and real-world experiments. The Franka robot is with unknown parameters of inertial matrix, gravitational forces, coriolis and centrifugal matrix, and utilizing ROS for module's communication. Therefore, we set \eqref{my_equ45} equal to zero in this section. We adapt existing VSGP code \cite{gomez2023adaptive} and customize kernel functions via Pyro \cite{bingham2019pyro}. Codes run on a 2.30GHz Core i7 CPU with 16GB RAM. For static obstacles, models are first trained and evaluated using data collected along the y-axis with the same motion scheme. After a predefined period, the evaluation is switched to the z-axis to assess the model's adaptability with unseen data, as depicted in Fig. \ref{fig3}. In this process, the uncertain component with HOCBF abbreviates as TUC. We optimize hyperparameters for the AFVSGP approach (ours) with only a 200-point dataset and 20 inducing points. By contrast, non-adaptive VSGP \cite{titsias2009variational}, conventional offline GP \cite{9683743}, and SOGP methods \cite{bui2017streaming} use a 2000-point dataset for one cycle offline training along y-axis. For dynamic obstacles, the end-effector moves cyclically along the y-axis, and its efficacy is compared to the SOGP method.

For static obstacles, Table I shows the mean squared error (MSE) and online training time (Tr. Time) across different methods, while Fig. \ref{fig2}(d) confirms AFVSGP's ability for real-time adaptive learning. In contrast, VSGP, depicted in Fig. \ref{fig2}(b), lacks a forgetting mechanism, which renders it susceptible to the influence of prior trajectory data, consequently biasing its real-time learning. The offline GP algorithm, presented in Fig. \ref{fig2}(a), inadequately adapts to new trajectory data, resulting in suboptimal data fitting post-trajectory change. Figs. \ref{fig2}(c) and \ref{fig2}(d) scrutinize the volume of data collected during the online learning process. Our algorithm's lower computational complexity enables real-time collection of a more extensive dataset, yielding higher data density than the SOGP method \cite{bui2017streaming}.

\begin{figure}[!t]
    \centering
    \begin{tikzpicture}
        \begin{scope}[xshift=1.11cm]
            \node[above right] (fig60) at (0,0){\includegraphics[width=0.45\linewidth,height=0.25\linewidth]{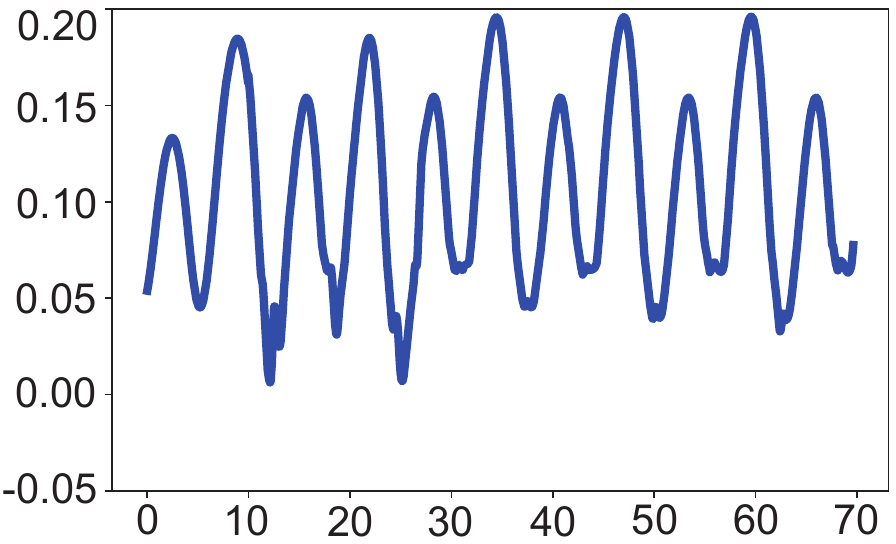}};
            \node[rectangle, rounded corners=1pt,minimum width=1cm,minimum height=0.3cm,draw=gray,fill=white!10](box) at ($(fig60.north east)+(-0.73,-0.36)$) {};
            \node (label) at ($(box.west)+(0.7,0.0)$) {\tiny $h(x)$};
            \draw[draw=blue,line width=1pt] ($(label.west)+(-0.25,-0.0)$) -- ($(label.west)+(0.05,-0.0)$);
            \node(text) at ($(fig60.south)+(1.3,+0.5)$) {\scriptsize $h(x)=0$};
            \draw[thick,-stealth,red] (text.west) -- ++(-0.3,0.2);
            \node at ($(fig60.south)+(0,-0.35)$) {\normalsize (b) AFVSGP};
            \node[rotate=90] at ($(fig60.west)+(-0,0.0)$) {\scriptsize Distance$[m]$};
            \node at ($(fig60.south)+(0,-0.02)$) {\scriptsize time$[s]$};
            \draw[draw=red, dashed, line width=0.5pt] ($(fig60.south west)+(0.68,0.72)$) -- ($(fig60.south east)+(-0.2,0.72)$);
        \end{scope}
        \begin{scope}[xshift=-3.2cm]
            \node[above right] (fig27) at (0,0){\includegraphics[width=0.45\linewidth,height=0.25\linewidth]{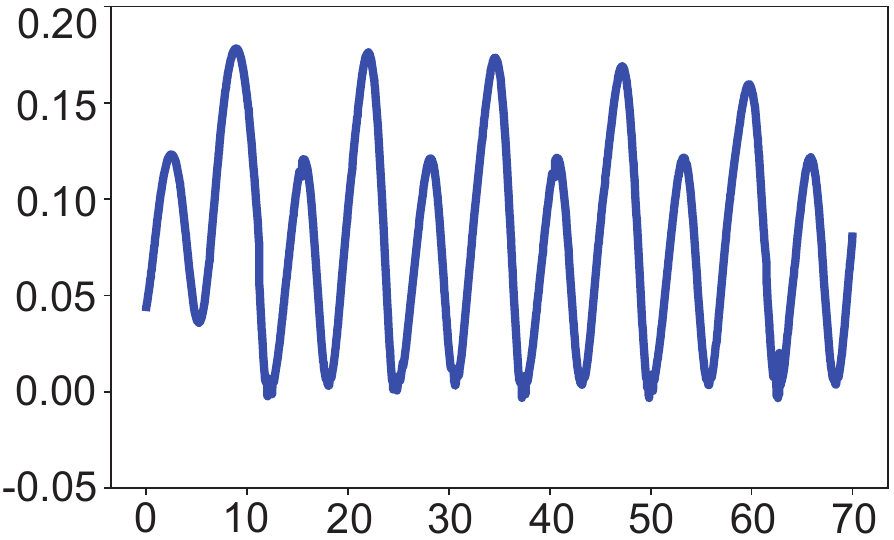}};
            \node[rectangle, rounded corners=1pt,minimum width=1cm,minimum height=0.3cm,draw=gray,fill=white!10](box) at ($(fig27.north east)+(-0.73,-0.36)$) {};
            \node (label) at ($(box.west)+(0.7,0.)$) {\tiny $h(x)$};
            \draw[draw=blue,line width=1pt] ($(label.west)+(-0.25,-0.0)$) -- ($(label.west)+(0.05,-0.0)$);
            \node(text) at ($(fig27.south)+(1.3,+0.5)$) {\scriptsize $h(x)=0$};
            \draw[thick,-stealth,red] (text.west) -- ++(-0.3,0.2);
            \node at ($(fig27.south)+(0,-0.35)$) {\normalsize (a) SOGP};
            \node[rotate=90] at ($(fig27.west)+(-0.,0)$) {\scriptsize Distance$[m]$};
            \node at ($(fig27.south)+(0,-0.02)$) {\scriptsize time$[s]$};
            \draw[draw=red, dashed, line width=0.5pt] ($(fig27.south west)+(0.68,0.72)$) -- ($(fig27.south east)+(-0.2,0.72)$);
        \end{scope}
    \end{tikzpicture}
    \caption{\quad  Control barrier function across different learning methods during dynamic obstacle avoidance. Here, $h(x)$ represents the minimum distance to the unsafe area's boundary. }
    \label{fig4}
    \vspace{-0.7em}
\end{figure}
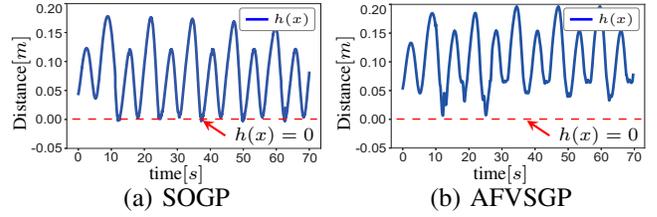

For dynamic obstacle avoidance, as depicted in Fig. \ref{fig2}(f), our algorithm achieves a close fit with real-time data. Conversely, the SOGP-based method, illustrated in Fig. \ref{fig2}(e), struggles to incorporate sufficient real-time data effectively due to prolonged training durations. Fig. \ref{fig4}(b) illustrates that our algorithm successfully maintains a positive CBF domain to ensure safety. In contrast, the SOGP methods, as shown in Fig. \ref{fig4}(a), result in a minimal distance between the end-effector and the unsafe area border that is less than zero, leading to frequent collisions.

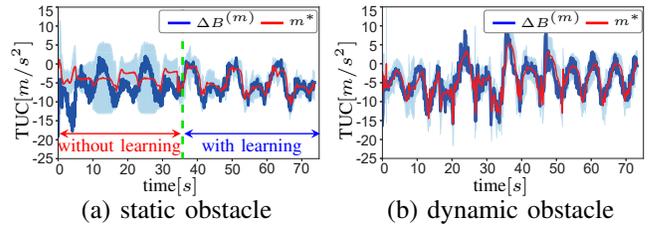
\begin{figure}[!t]
  \centering
  \hspace{0em}
  \begin{tikzpicture}
    \begin{scope}[xshift=-3cm]
        \node[above right] (fig63) at (0,0){\includegraphics[ height=0.265\linewidth]{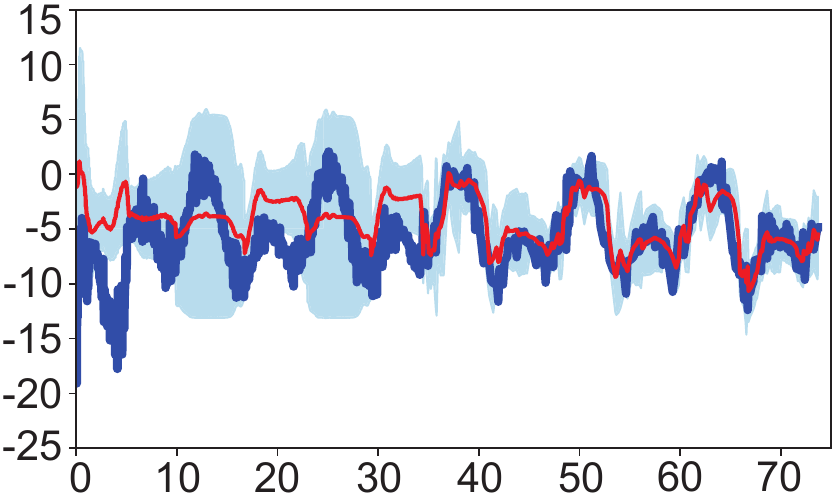}};
        \node at ($(fig63.south)+(0,-0.35)$) {\normalsize (a) static obstacle};
        \draw[draw=green!90!black,dashed,line width=1pt] ($(fig63.north)+(0.08,-0.2)$) -- ($(fig63.south)+(0.08,0.35)$);

        \node[rectangle, rounded corners=1pt,minimum width=2.1cm,minimum height=0.3cm,draw=gray,fill=white!10](box) at ($(fig63.north east)+(-1.23,-0.36)$) {};
            \node (label) at ($(box.west)+(0.9,0.03)$) {\tiny $\Delta B^{(m)}$};
            \node (label2) at ($(box.east)+(-0.2,0.03)$) {\tiny $m^*$};
            \draw[draw=blue,line width=1pt] ($(label.west)+(-0.25,-0.05)$) -- ($(label.west)+(0.05,-0.05)$) node (legend){};
            \draw[draw=red,line width=1pt] ($(label2.west)+(-0.25,-0.05)$) -- ($(label2.west)+(0.05,-0.05)$);
            \node[rotate=90] at ($(fig63.west)+(0,0)$) {\scriptsize TUC$[m\slash s^2]$};
            \node at ($(fig63.south)+(-0.08,-0.02)$) {\scriptsize time$[s]$};
            \node(text2) at ($(fig63.south)+(-0.76,0.5)$) {\scriptsize \textcolor{red}{ without learning} };
            \draw[stealth-stealth, red, line width=0.5pt]($(text2.north west)+(0.15,-0.05)$) -- ($(text2.north east)+(-0.13,-0.05)$);
            \node(text3) at ($(fig63.south)+(1,0.5)$) {\scriptsize \textcolor{blue}{ with learning}};
            \draw[stealth-stealth, blue, line width=0.5pt]($(text3.north west)+(-0.11,-0.05)$) -- ($(text3.north east)+(0.12,-0.05)$);
    \end{scope}

    \begin{scope}[xshift=1.3cm]
        \node[above right] (fig64) at (0,0){\includegraphics[ height=0.265\linewidth]{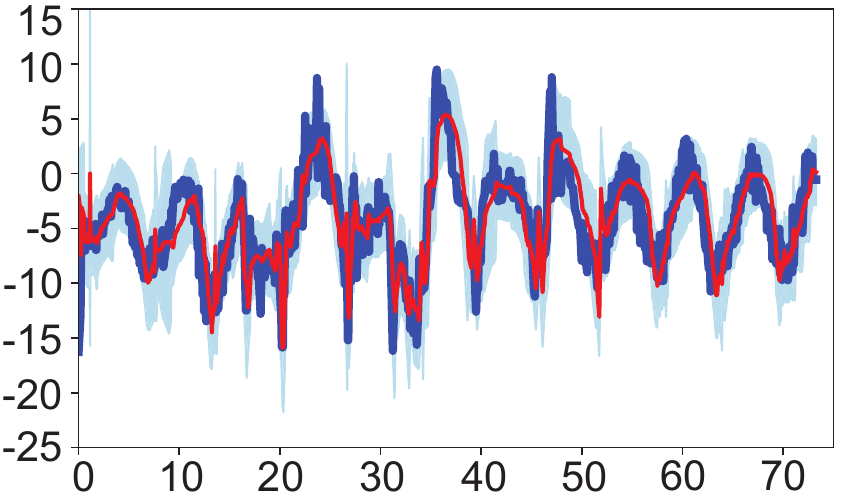}};
        \node at ($(fig64.south)+(0,-0.35)$) {\normalsize (b) dynamic obstacle};
        \node[rectangle, rounded corners=1pt,minimum width=2.1cm,minimum height=0.3cm,draw=gray,fill=white!10](box) at ($(fig64.north east)+(-1.23,-0.36)$) {};
            \node (label) at ($(box.west)+(0.9,0.03)$) {\tiny $\Delta B^{(m)}$};
            \node (label2) at ($(box.east)+(-0.2,0.03)$) {\tiny $m^*$};
            \draw[draw=blue,line width=1pt] ($(label.west)+(-0.25,-0.05)$) -- ($(label.west)+(0.05,-0.05)$) node (legend){};
            \draw[draw=red,line width=1pt] ($(label2.west)+(-0.25,-0.05)$) -- ($(label2.west)+(0.05,-0.05)$);
            \node[rotate=90] at ($(fig64.west)+(-0.0,0)$) {\scriptsize TUC$[m\slash s^2]$};
            \node at ($(fig64.south)+(0,-0.02)$) {\scriptsize time$[s]$};
    \end{scope}
  \end{tikzpicture}
  \caption{\quad Online learning outcomes for real-world obstacle avoidance with the Franka robot using the AFVSGP-HOCBF method.}
  \label{fig5}
  \vspace{-0.4em}
\end{figure}

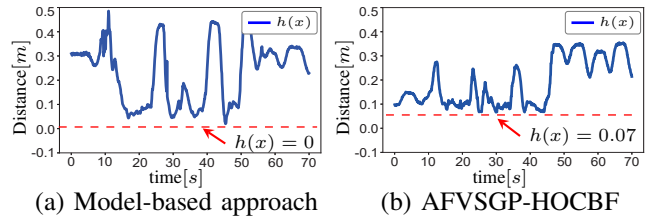
\begin{figure}[!t]
  \centering
    \begin{tikzpicture}
        \begin{scope}[xshift=-3cm]
            \node[above right] (fig69) at (0,0){\includegraphics[width=0.45\linewidth,height=0.25\linewidth]{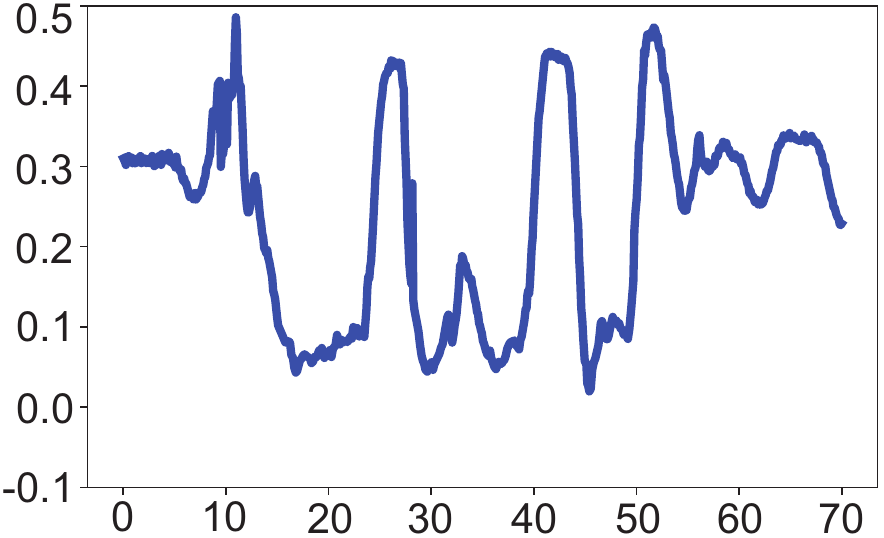}};
            \node[rectangle, rounded corners=1pt,minimum width=1cm,minimum height=0.3cm,draw=gray,fill=white!10](box) at ($(fig69.north east)+(-0.7,-0.35)$) {};
            \node (label) at ($(box.west)+(0.7,0.0)$) {\tiny $h(x)$};
            \draw[draw=blue,line width=1pt] ($(label.west)+(-0.25,-0.0)$) -- ($(label.west)+(0.05,-0.0)$);
            \node(text) at ($(fig69.south)+(1.3,+0.45)$) {\scriptsize $h(x)=0$};
            \draw[thick,-stealth,red] (text.west) -- ++(-0.3,0.2);
            \node at ($(fig69.south)+(0,-0.35)$) {\normalsize (a) Model-based approach};
            \node[rotate=90] at ($(fig69.west)+(-0.,0)$) {\scriptsize Distance$[m]$};
            \node at ($(fig69.south)+(0,-0.02)$) {\scriptsize time$[s]$};
            \draw[draw=red, dashed, line width=0.5pt] ($(fig69.south west)+(0.5,0.67)$) -- ($(fig69.south east)+(-0.2,0.67)$);

        \end{scope}
        \begin{scope}[xshift=1.3cm]
            \node[above right] (fig70) at (0,0){\includegraphics[width=0.45\linewidth,height=0.25\linewidth]{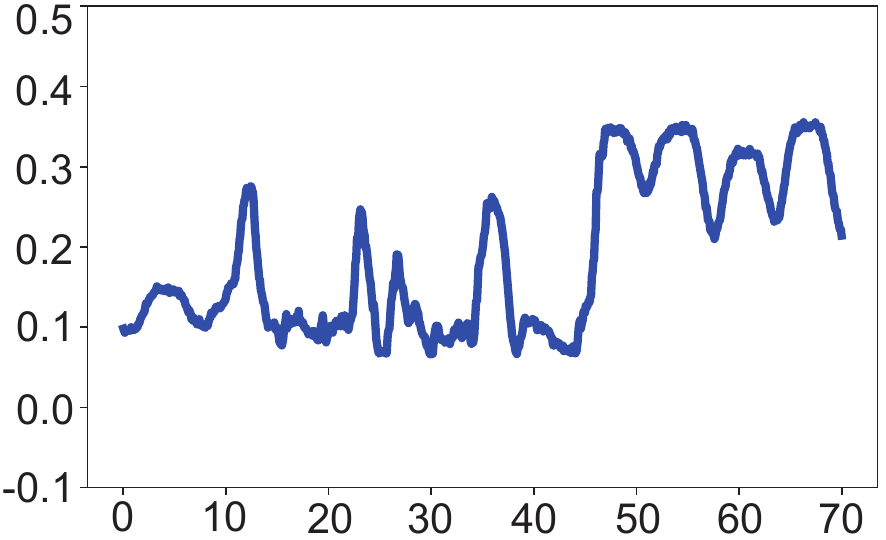}};
            \node[rectangle, rounded corners=1pt,minimum width=1cm,minimum height=0.3cm,draw=gray,fill=white!10](box) at ($(fig70.north east)+(-0.7,-0.35)$) {};
            \node (label) at ($(box.west)+(0.7,0.)$) {\tiny $h(x)$};
            \draw[draw=blue,line width=1pt] ($(label.west)+(-0.25,-0.0)$) -- ($(label.west)+(0.05,-0.0)$);
            \node(text) at ($(fig70.south)+(1.1,+0.55)$) {\scriptsize $h(x)=0.07$};
            \draw[thick,-stealth,red] (text.west) -- ++(-0.3,0.23);
            \node at ($(fig70.south)+(0,-0.35)$) {\normalsize (b) AFVSGP-HOCBF};
            \node[rotate=90] at ($(fig70.west)+(-0.,0)$) {\scriptsize Distance$[m]$};
            \node at ($(fig70.south)+(0,-0.02)$) {\scriptsize time$[s]$};
            \draw[draw=red, dashed, line width=0.5pt] ($(fig70.south west)+(0.55,0.83)$) -- ($(fig70.south east)+(-0.25,0.83)$);
        \end{scope}
    \end{tikzpicture}
  \caption{\quad Comparison of control barrier function between known-parameter methods and our AFVSGP-HOCBF approach with fully unknown parameters in real-world tests.}
  \label{fig6}
  \vspace{-1.2em}
\end{figure}

In the real-world experiment, the experimental platform utilizes a Franka robot equipped with two Intel D415 depth cameras for spatial localization. Obstacle classification is performed using the YOLOv8n model \cite{Jocher2023}, as shown in Fig. \ref{fig1}. To safeguard the robotic manipulator and personnel, the experiments exclude failed simulation methods, focusing solely on known-parameter methods and AFVSGP-HOCBF (ours) with fully unknown parameters. The experimental design is segmented into three parts. The first part involves a comparative analysis of data fitting with and without learning in the presence of static obstacles. The second part focuses on complete online learning results under dynamic obstacle conditions. Finally, the third part contrasts the performance with known-parameter approaches. Following an abrupt switch to our algorithm, as shown in Fig. \ref{fig5}(a), the proposed method rapidly adapts to the data in the presence of static obstacles. In Fig. \ref{fig5}(b), due to its rapid inference capabilities, the proposed algorithm also effectively fits the required data in the presence of dynamic obstacles. In Fig. \ref{fig6}(b), the data shows a distance greater than 0.07m from the origin, indicating that the minimal distance between end-effector and the border line of unsafe area is 0.07m. This behavior is little conservative compared to model-based approaches, as shown in Fig. \ref{fig6}(a), a result of the confidence intervals inherent to our method. Nonetheless, our method ensures safety while maintaining an acceptable level of conservatism.

\section{Conclusion}
This paper introduces a novel adaptive online learning framework to address real-time safety concerns in higher-order nonlinear systems operating under model uncertainty and in non-stable environments. In the process of designing rules to select the most representative data points and reconstructing GPs model, the time complexity for updating prediction has been reduced from $\mathcal{O}\left(NM^2\right)$ to $\mathcal{O}\left(M^3\right)$. This ensures real-time applicability in high-frequency systems. The feasibility of the algorithm in real-world scenarios is substantiated through empirical experiments. For future work, we will integrate additional factors into the proposed framework based on CBF, which is explored to address the challenges posed by multi-obstacle avoidance scenarios.

\balance
\bibliographystyle{ieeetr}
\bibliography{ref}

\begin{thebibliography}{10}

\bibitem{8796030}
A.~D. Ames, S.~Coogan, M.~Egerstedt, G.~Notomista, K.~Sreenath, and P.~Tabuada,
  ``Control barrier functions: Theory and applications,'' in {\em 2019 18th
  European Control Conference (ECC)}, pp.~3420--3431, 2019.

\bibitem{nguyen2015safety}
Q.~Nguyen and K.~Sreenath, ``Safety-critical control for dynamical bipedal
  walking with precise footstep placement,'' {\em IFAC-PapersOnLine}, vol.~48,
  no.~27, pp.~147--154, 2015.

\bibitem{9029446}
T.~D. Son and Q.~Nguyen, ``Safety-critical control for non-affine nonlinear
  systems with application on autonomous vehicle,'' in {\em 2019 IEEE 58th
  Conference on Decision and Control (CDC)}, pp.~7623--7628, 2019.

\bibitem{10132404}
Z.~Bing, L.~Knak, L.~Cheng, F.~O. Morin, K.~Huang, and A.~Knoll,
  ``Meta-reinforcement learning in nonstationary and nonparametric
  environments,'' {\em IEEE Transactions on Neural Networks and Learning
  Systems}, pp.~1--15, 2023.

\bibitem{8460471}
L.~Wang, E.~A. Theodorou, and M.~Egerstedt, ``Safe learning of quadrotor
  dynamics using barrier certificates,'' in {\em 2018 IEEE International
  Conference on Robotics and Automation (ICRA)}, pp.~2460--2465, 2018.

\bibitem{9112342}
Y.~Chen, A.~Singletary, and A.~D. Ames, ``Guaranteed obstacle avoidance for
  multi-robot operations with limited actuation: A control barrier function
  approach,'' {\em IEEE Control Systems Letters}, vol.~5, no.~1, pp.~127--132,
  2021.

\bibitem{10341769}
X.~Yao, Z.~Bing, G.~Zhuang, K.~Chen, H.~Zhou, K.~Huang, and A.~Knoll,
  ``Learning from symmetry: Meta-reinforcement learning with symmetrical
  behaviors and language instructions,'' in {\em 2023 IEEE/RSJ International
  Conference on Intelligent Robots and Systems (IROS)}, pp.~5574--5581, 2023.

\bibitem{9147463}
A.~J. Taylor and A.~D. Ames, ``Adaptive safety with control barrier
  functions,'' in {\em 2020 American Control Conference (ACC)}, pp.~1399--1405,
  2020.

\bibitem{9353988}
Q.~Nguyen and K.~Sreenath, ``Robust safety-critical control for dynamic
  robotics,'' {\em IEEE Transactions on Automatic Control}, vol.~67, no.~3,
  pp.~1073--1088, 2022.

\bibitem{9129764}
B.~T. Lopez, J.-J.~E. Slotine, and J.~P. How, ``Robust adaptive control barrier
  functions: An adaptive and data-driven approach to safety,'' {\em IEEE
  Control Systems Letters}, vol.~5, no.~3, pp.~1031--1036, 2021.

\bibitem{dong2023novel}
J.~Dong, W.~Si, and C.~Yang, ``A novel human-robot skill transfer method for
  contact-rich manipulation task,'' {\em Robotic Intelligence and Automation},
  vol.~43, no.~3, pp.~327--337, 2023.

\bibitem{9772990}
Z.~Bing, H.~Zhou, R.~Li, X.~Su, F.~O. Morin, K.~Huang, and A.~Knoll, ``Solving
  robotic manipulation with sparse reward reinforcement learning via
  graph-based diversity and proximity,'' {\em IEEE Transactions on Industrial
  Electronics}, vol.~70, no.~3, pp.~2759--2769, 2023.

\bibitem{beckers2017stable}
T.~Beckers, J.~Umlauft, and S.~Hirche, ``Stable model-based control with
  gaussian process regression for robot manipulators,'' {\em
  IFAC-PapersOnLine}, vol.~50, no.~1, pp.~3877--3884, 2017.

\bibitem{10160626}
Z.~Bing, A.~Koch, X.~Yao, K.~Huang, and A.~Knoll, ``Meta-reinforcement learning
  via language instructions,'' in {\em 2023 IEEE International Conference on
  Robotics and Automation (ICRA)}, pp.~5985--5991, 2023.

\bibitem{9466373}
Z.~Bing, M.~Brucker, F.~O. Morin, R.~Li, X.~Su, K.~Huang, and A.~Knoll,
  ``Complex robotic manipulation via graph-based hindsight goal generation,''
  {\em IEEE Transactions on Neural Networks and Learning Systems}, vol.~33,
  no.~12, pp.~7863--7876, 2022.

\bibitem{kocijan2016modelling}
J.~Kocijan, {\em Modelling and control of dynamic systems using Gaussian
  process models}.
\newblock Springer, 2016.

\bibitem{9303847}
P.~Jagtap, G.~J. Pappas, and M.~Zamani, ``Control barrier functions for unknown
  nonlinear systems using gaussian processes,'' in {\em 2020 59th IEEE
  Conference on Decision and Control (CDC)}, pp.~3699--3704, 2020.

\bibitem{9196709}
D.~D. Fan, J.~Nguyen, R.~Thakker, N.~Alatur, A.-a. Agha-mohammadi, and E.~A.
  Theodorou, ``Bayesian learning-based adaptive control for safety critical
  systems,'' in {\em 2020 IEEE International Conference on Robotics and
  Automation (ICRA)}, pp.~4093--4099, 2020.

\bibitem{gal2016dropout}
Y.~Gal and Z.~Ghahramani, ``Dropout as a bayesian approximation: Representing
  model uncertainty in deep learning,'' in {\em international conference on
  machine learning}, pp.~1050--1059, PMLR, 2016.

\bibitem{harrison2020meta}
J.~Harrison, A.~Sharma, and M.~Pavone, ``Meta-learning priors for efficient
  online bayesian regression,'' in {\em Algorithmic Foundations of Robotics
  XIII: Proceedings of the 13th Workshop on the Algorithmic Foundations of
  Robotics 13}, pp.~318--337, Springer, 2020.

\bibitem{bui2017streaming}
T.~D. Bui, C.~Nguyen, and R.~E. Turner, ``Streaming sparse gaussian process
  approximations,'' {\em Advances in Neural Information Processing Systems},
  vol.~30, 2017.

\bibitem{9658123}
V.~Dhiman, M.~J. Khojasteh, M.~Franceschetti, and N.~Atanasov, ``Control
  barriers in bayesian learning of system dynamics,'' {\em IEEE Transactions on
  Automatic Control}, vol.~68, no.~1, pp.~214--229, 2023.

\bibitem{9683743}
F.~Castañeda, J.~J. Choi, B.~Zhang, C.~J. Tomlin, and K.~Sreenath, ``Pointwise
  feasibility of gaussian process-based safety-critical control under model
  uncertainty,'' in {\em 2021 60th IEEE Conference on Decision and Control
  (CDC)}, pp.~6762--6769, 2021.

\bibitem{xu2015robustness}
X.~Xu, P.~Tabuada, J.~W. Grizzle, and A.~D. Ames, ``Robustness of control
  barrier functions for safety critical control,'' {\em IFAC-PapersOnLine},
  vol.~48, no.~27, pp.~54--61, 2015.

\bibitem{titsias2009variational}
M.~Titsias, ``Variational learning of inducing variables in sparse gaussian
  processes,'' in {\em Artificial intelligence and statistics}, pp.~567--574,
  PMLR, 2009.

\bibitem{castaneda2021gaussian}
F.~Castaneda, J.~J. Choi, B.~Zhang, C.~J. Tomlin, and K.~Sreenath, ``Gaussian
  process-based min-norm stabilizing controller for control-affine systems with
  uncertain input effects and dynamics,'' in {\em 2021 American Control
  Conference (ACC)}, pp.~3683--3690, IEEE, 2021.

\bibitem{gomez2023adaptive}
V.~G{\'o}mez-Verdejo and M.~Mart{\'\i}nez-Ram{\'o}n, ``Adaptive sparse gaussian
  process,'' {\em arXiv preprint arXiv:2302.10325}, 2023.

\bibitem{csato2002sparse}
L.~Csat{\'o} and M.~Opper, ``Sparse on-line gaussian processes,'' {\em Neural
  computation}, vol.~14, no.~3, pp.~641--668, 2002.

\bibitem{9961103}
Y.~He and Y.~Zhao, ``Adaptive robust control of uncertain euler–lagrange
  systems using gaussian processes,'' {\em IEEE Transactions on Neural Networks
  and Learning Systems}, 2022, DOI: 10.1109/TNNLS.2022.3222405.

\bibitem{wendland2004scattered}
H.~Wendland, {\em Scattered data approximation}, vol.~17.
\newblock Cambridge university press, 2004.

\bibitem{srinivas2012information}
N.~Srinivas, A.~Krause, S.~M. Kakade, and M.~W. Seeger, ``Information-theoretic
  regret bounds for gaussian process optimization in the bandit setting,'' {\em
  IEEE transactions on information theory}, vol.~58, no.~5, pp.~3250--3265,
  2012.

\bibitem{dean2021guaranteeing}
S.~Dean, A.~Taylor, R.~Cosner, B.~Recht, and A.~Ames, ``Guaranteeing safety of
  learned perception modules via measurement-robust control barrier
  functions,'' in {\em Conference on Robot Learning}, pp.~654--670, PMLR, 2021.

\bibitem{buch2021robust}
J.~Buch, S.-C. Liao, and P.~Seiler, ``Robust control barrier functions with
  sector-bounded uncertainties,'' {\em IEEE Control Systems Letters}, vol.~6,
  pp.~1994--1999, 2021.

\bibitem{bingham2019pyro}
E.~Bingham, J.~P. Chen, M.~Jankowiak, F.~Obermeyer, N.~Pradhan, T.~Karaletsos,
  R.~Singh, P.~Szerlip, P.~Horsfall, and N.~D. Goodman, ``Pyro: Deep universal
  probabilistic programming,'' {\em The Journal of Machine Learning Research},
  vol.~20, no.~1, pp.~973--978, 2019.

\bibitem{Jocher2023}
G.~Jocher, A.~Chaurasia, and J.~Qiu, ``{YOLO by Ultralytics},'' 2023.
\newblock Accessed: August 10, 2023.

\end{thebibliography}

\end{document}